\documentclass[10pt,journal]{IEEEtran}
\usepackage{latexsym,amssymb,amsfonts,enumerate,graphicx,subfigure}
\usepackage{amsthm,amsmath}
\usepackage{booktabs}
\usepackage{multirow,color,ulem}
\usepackage{algorithm,algorithmic}
\usepackage{cite}
\usepackage{tablefootnote,footnote}
\usepackage[colorlinks,linkcolor=red,urlcolor=blue,anchorcolor=blue,citecolor=green]{hyperref}
\usepackage{epstopdf}
\newcommand{\no}{\nonumber}
\newcommand{\be}{\begin{eqnarray}}
\newcommand{\ben}{\begin{eqnarray*}}
\newcommand{\en}{\end{eqnarray}}
\newcommand{\enn}{\end{eqnarray*}}
\newcommand{\vep}{\varepsilon}

\newtheorem{theorem}{Theorem}

\graphicspath{{draw/}}
\begin{document}

\title{Fast Density Estimation for Density-based Clustering Methods}

%
\author{
Difei~Cheng\thanks{D. cheng, R. Xu and R. Jin are with Academy of Mathematics and Systems Science,
	Chinese Academy of Sciences, Beijing 10090, China and School of Mathematical Sciences,
	University of Chinese Academy of Sciences, Beijing 10049, China}, {\it Student~Member,~IEEE},
Ruihang~Xu, {\it Student~Member,~IEEE},
Bo Zhang\thanks{B. Zhang is with LSEC and Academy of Mathematics and Systems Science,
Chinese Academy of Sciences, Beijing 100190, China and School of Mathematical Sciences,
University of Chinese Academy of Sciences, Beijing 100049, China (email: b.zhang@amt.ac.cn)
\bf Corresponding author: Bo Zhang}, {\it Member, IEEE,}\\
and Ruinan~Jin, {\it Student~Member,~IEEE},
}

\maketitle

\begin{abstract}
Density-based clustering algorithms are widely used for discovering clusters in pattern recognition
and machine learning. They can deal with non-hyperspherical clusters and are robust to
outliers. However, the runtime of density-based algorithms is heavily dominated by neighborhood finding
and density estimation which is time-consuming. Meanwhile, the traditional acceleration methods
using indexing techniques such as KD trees may not be effective in dealing with high-dimensional data.
To address these issues, this paper proposes a fast range query algorithm, called Fast Principal
Component Analysis Pruning (FPCAP), with the help of the fast principal component analysis technique
in conjunction with geometric information provided by the principal attributes of the data.
FPCAP can deal with high-dimensional data and can be easily applied to density-based methods to prune
unnecessary distance calculations in neighborhood finding and density estimation.
As an application in density-based clustering methods, FPCAP is combined with the Density Based Spatial
Clustering of Applications with Noise (DBSCAN) algorithm, and an improved DBSCAN (called IDBSCAN) is then
obtained. IDBSCAN preserves the advantage of DBSCAN and, meanwhile, greatly reduces the computation
of redundant distances.
Experiments on seven benchmark datasets demonstrate that the proposed algorithm improves the
computational efficiency significantly.
\end{abstract}




\IEEEpeerreviewmaketitle

\section{Introduction}\label{sec:introduction}

\IEEEPARstart{C}{lustering} aims to partition a set of data objects into several clusters so that objects
in each cluster keep high degree of similarity, and thus has been widely used in data
mining \cite{berkhin2006survey}, vector quantization \cite{coates2011importance}, dimension
reduction \cite{boutsidis2014randomized} and manifold learning \cite{canas2012learning}.

Density-based clustering is one of the most important clustering methods. It is based on density
estimation of data points and defines clusters as dense regions separated by low-density regions.
It does not need to know the number of clusters, and meanwhile, it has the ability to discover clusters
with arbitrary shapes and is robust to outliers.
So far, many density-based methods have been proposed, such as DBSCAN \cite{ester1996density},
the Ordering Points To Identify the Clustering Structure (OPTICS) \cite{ankerst1999optics},
Clustering by Fast Search and Find of Density Peaks (CFSFDP) \cite{rodriguez2014clustering}
and Mean-Shift Clustering \cite{cheng1995mean,anand2013semi}.

DBSCAN is probably the most prominent density-based clustering algorithm so far. It is based on
the key idea that for each core data point of a cluster its neighborhood of a given radius $\vep$
has to contain at least a minimum number of data points ($\textrm{MinPts}$). However, DBSCAN has some
disadvantages. First, the parameters $\vep$ and $\textrm{MinPts}$ have a significant influence
on the clustering results and are difficult to choose. OPTICS \cite{ankerst1999optics} was then proposed
to overcome this difficulty by creating an augmented ordering of the database representing
a density-based clustering structure which includes the information of the clustering results obtained
by DBSCAN corresponding to a broad range of parameters settings.
Secondly, the time complexity of DBSCAN is $O(n^2)$. Thus many improved methods have been proposed to
accelerate DBSCAN, which can be roughly divided into two categories: sampling-based improvements and
partition-based improvements.
Sampling-based methods \cite{chen2021block,borah2004improved,liu2006fast,jang2019dbscan++,chen2018fast}
try to reduce the number of range queries by skipping some queries from specific points in the process of
finding neighbors and calculating the density of each point,
whilst partition-based methods \cite{mahran2008using,gunawan2013faster,gan2017hardness} use grid partitions
to divide the data into several groups to process separately.
When finding neighbors, most of the above improved DBSCAN algorithms speed up DBSCAN by reducing
the number of range queries. But some sampling-based methods, such as those in \cite{chen2021block}
and \cite{chen2018fast}, still need to calculate the ${\vep}/2$-neighborhoods and $2\vep$-neighborhoods
many times. Thus it is important to reduce the range query time during neighborhood finding.
Traditionally, indexing techniques such as KD-tree \cite{bentley1977complexity}, Cover
tree \cite{beygelzimer2006cover}, quadtree-like hierarchical tree \cite{beygelzimer2006cover}
and R tree \cite{guttman1984r} are used to reduce the range query time in these sampling-based algorithms.
However, the construction of the tree structures is complex, and these indexing techniques such as KD tree
are very difficult to deal with high-dimensional data sets.

Motivated by the work in \cite{lai2010fast} which applies the Fast Principal Component Analysis (Fast-PCA)
technique \cite{sharma2007fast} in accelerating the global $k$-means algorithm \cite{likas2003global},
we propose a fast range queries method (called FPCAP), based on the Fast-PCA technique in conjunction
with certain geometric information provided by the principal attributes of the data.
FPCAP can greatly reduce the range query time by reducing the computation of redundant distances
during neighborhood finding and density estimation, and meanwhile, it avoids any complex data structures
which are needed to embed the data points themselves and are not easy to implement.
Further, FPCAP can deal with high-dimensional data sets.
In experiments, FPCAP is compared with the KD tree indexing technique.
As an application in density-based clustering methods, FPCAP is applied to DBSCAN to obtain an
improved DBSCAN algorithm which is called IDBSCAN.
IDBSCAN is compared with the original DBSCAN algorithm as well as
the original DBSCAN algorithm with the KD tree indexing technique.
The experimental results on seven clustering benchmark datasets illustrate that both FPCAP and IDBSCAN
outperform the other compared algorithms on computational efficiency. In addition, IDBSCAN is an exact DBSCAN
algorithm which produces the same results as did by DBSCAN.

The remaining part of the paper is organized as follows.
In Section \ref{sec2}, we briefly introduce some related work, the DBSCAN algorithm,
the Fast-PCA algorithm and the KD-tree algorithm.
Our proposed algorithm is proposed in Section \ref{sec3}.
Experimental results are provided in Section \ref{sec4}, and conclusions are given in Section \ref{sec5}.

\section{Related Work}\label{sec2}

\subsection{The DBSCAN algorithm}

DBSCAN \cite{ester1996density} has two important parameters $\varepsilon$ and $\textrm{MinPts}$.
A point $x$ is called a core point if the number of points within the $\varepsilon$-neighborhood
of $x$ is more than $\textrm{MinPts}$. A point $y$ is called directly density-reachable from a core point $x$
if $y$ is in the $\varepsilon$-neighborhood of $x$. A point $y$ is called density-reachable from a
core point $x$ if there is a series of points $y_1,y_2,...,y_n$ with $y=y_n$, $x=y_1$ and for
any pair $y_i,y_{i+1}$ we have that $y_{i+1}$ is directly density-reachable from $y_i$.
Two points are density-connected if they both are density-reachable from the same core point.
A cluster $C$ in DBSCAN satisfies the following two conditions:
\begin{itemize}
\item Maximality: any point which is density-reachable from a core point in cluster $C$ is also
in cluster $C$.
\item Connectivity: any two points in cluster $C$ are density-connected.
\end{itemize}

Details of DBSCAN are described in Algorithm \ref{alg:DBSCAN}.
\begin{algorithm}
	\caption{The DBSCAN algorithm}\label{alg:DBSCAN}
	\hspace*{0.02in}{\bf Input:} $D=\{x_{1},x_{2},\ldots,x_{n}\}$, $\varepsilon$, $\textrm{MinPts}$ \\
	\hspace*{0.02in}{\bf Output:} $classifications$
	\begin{algorithmic}[1]
		\STATE Initialize $classifications$ to an $n$-dimensional vector with its component being Unclassified;
               $seed=\emptyset$; $cluster\_{id}=1$; $ID=1$; $id=1$
		\STATE Calculate
               $N_{\vep}(x_{id})=\{x\in D|d(x,x_{id})<\vep\}$, where $x_{id}\in D$ and $d(x,y)$
               denotes the distance between two points. 
		\STATE If $|N_{\vep}(x_{id})|<\textrm{MinPts}$ and $x_{id}$'s classification is Unclassified, then $classifications(id)$ is Noise,
               where $|D|$ denotes the cardinality of the set $D$;
		       otherwise, for every point $y$ whose classification is Unclassified or Noise
               in $N_{\vep}(x_{id})$, we set
		\ben
			classifications(p):=cluster\_{id},
		\enn
        where $p$ is index of $y$, and if $y$'s classification is Unclassified, update $seed=seed\cup\{y\}$.
		\STATE If $seed\neq\emptyset$, update $seed=seed/x_{id}$
		\STATE  If $seed\neq\emptyset$, choose a point $y$ in $seed$, update $id=index\;of\;y$ and go back
               to Step 2. Otherwise, if $classification(id)=cluster\_{id}$, update $cluster\_{id}=cluster\_{id}+1$
		\STATE Update $ID=ID+1$, $id=ID$. If $classifications(id)$ is Unclassified, go back to Step 2;
               otherwise, go back to Step 6 until $ID=n$.	
	\end{algorithmic}
\end{algorithm}

\subsection{Fast-PCA}

Fast-PCA \cite{sharma2007fast} is designed to find $h$ leading eigenvectors by using a
fixed-point algorithm \cite{hyvarinen1997fast}. Its computational cost is much less than that of
the eigenvalue decomposition (EVD) based PCA. Algorithm \ref{alg:fpca} presents details of Fast-PCA.

\begin{algorithm}
	\caption{Fast-PCA}\label{alg:fpca}
    {\bf Input:} Data $X=[X_{1},\ldots,X_{h}]$ of size $n\times h$, where $X_{i}$ is the $i$th feature
    with size $n\times 1$ and $\sum X_{i}=0$, that is, $X$ is assumed to be zero-centered, and the desired
    number $h$ of leading eigenvectors. \\
    {\bf Output:} projection of data, $Z$, with size $n\times h$.\\
    \vspace*{-0.2in}
	\begin{algorithmic}[1]
		\STATE Compute covariance $\sum_{X}=XX^{T}$, and set $p=1$
		\STATE Randomly initialize eigenvector $\phi_{p}$ of size $n\times 1$
		\STATE Update $\phi_{p}$ as $\phi_{p}\gets\sum_{X}\phi_p$
		\STATE Do the Gram-Schmidt orthogonalization process for $\phi_p$:
		\begin{equation}
			\phi_p\gets\phi_p - \sum_{j=1}^{p-1}(\phi_p^{T}\phi_j)\phi_j
		\end{equation}
		\STATE Normalize $\phi_p$ by dividing it by its norm:
		\begin{equation}
			\phi_p\gets\phi_p/\|\phi_p\|
		\end{equation}
		\STATE If $\phi_p$ has not converged, go back to Step 3. (The Fast-PCA algorithm for the $p$-th
               basis vector converges when the new and old values $\phi_{p}$ point in the same direction,
               i.e. $\left(\phi_p^+\right)^T\phi_p\approx 1$, where $\phi_p^+$ is the new value of $\phi_p$).
		\STATE Set $p=p+1$ and go to Step 2 until $p=h$
		\STATE Get the orthogonal projection matrix: $\Phi=[\phi_1,\phi_2,...\phi_{h}]$.
               The columns of the projection matrix are sorted by the descending order of
               the corresponding eigenvalues.
		\STATE Get the projection of data: $Z=X\Phi$
	\end{algorithmic}
\end{algorithm}

\subsection{KD-tree}

KD-tree was introduced by Bentley \cite{bentley1977complexity,bentley1975multidimensional}
as a binary tree that stores $k$-dimensional data.
As a famous space-partitioning data structure used to organize points in a $k$-dimensional space,
KD-tree subdivides data like a binary tree at each recursive level of the tree
but uses $k$ keys for all levels of the tree which is different from a binary tree that uses a single key.
For example, to build a KD-tree from three-dimensional points in the $(x,y,z)$ coordinates,
the coordinate of keys would be chose circularly from $x,y,z$ for successive levels of the KD-tree.
A often-used scheme for cycling the keys chooses the coordinate that has the widest dispersion or
largest variance to be the key for a particular level of recursion, and the splitting node is
positioned at the spatial median of the chose coordinate.
As an acceleration structure, it has been used in a variety of applications, including range queries
for fixed-radius near neighbors and nearest neighbors searching. For example, the KD-tree indexing technique
is frequently used for the range query in the process of finding fixed-radius near neighbors or
calculating densities in density-based clustering. When the KD-tree is used for range queries, it first
locates the smallest sample subspace according to the judgment method of a binary tree and then
backtracks each parent node. When the distance between the target point and the parent node is
less than the threshold, the searching enters another subspace of the parent node.

\section{The proposed algorithm}\label{sec3}

\subsection{Fast-PCA pruning}

For a given data set $D$ of $n$ points, it is time-consuming to find neighbors for each point by
simply comparing the distances between the point and all other $n-1$ points in the data set, which
has a time complexity of $O(n^{2}h)$, where $h$ is the dimension of the data.
An indexing technique can be used to accelerate the above process of finding neighbors, but it is
difficult to deal with high-dimensional data.
To address this issue, we propose a fast range query algorithm to reduce the redundant distance
calculations and improve the computational efficiency, based on Fast-PCA in conjunction with some
geometric information provided by the principal attributes of the data.

Let $D=\{x_{1},x_{2},\ldots,x_{n}\}$ be a set of data points, where $x_{i}$ is an $h$-dimensional
vector and can be represented as $x_{i}=\sum_{j=1}^{h}x_{i,j}e_j$ with the $e_j$'s being
the $h$-dimensional orthonormal basis vectors.
Take $X=[x_{1}^{T},x_{2}^{T},\ldots,x_{n}^{T}]$ in Algorithm \ref{alg:fpca} and
denote by $\phi_{i}$ the eigenvector corresponding to the $i$th largest eigenvalue, $i=1,\ldots,h$,
obtained by Algorithm \ref{alg:fpca}.
Then $x_{i}$ can be rewritten as $x_{i}=\sum_{j=1}^{h}z_{i,j}\phi_{j}$,
where $z_{i,j}=\langle x_i,\phi_j\rangle$. In fact, $Z=(z_{i,j})_{n\times h}$ which is the projection
of data obtained by Algorithm \ref{alg:fpca}. For any $x_i\in D$ define
\be\label{x-split}
x_{i}:=\hat{z}_{i}+\tilde{z}_{i},
\en
where $\hat{z}_{i}=\sum_{j=1}^{h_1}z_{i,j}\phi_{j}$, $\tilde{z}_{i}=\sum_{j=h_{1}+1}^{h}z_{i,j}\phi_{j}$
and $h_1$ is an integer which is less than $h$ and to be determined later. Then
\be\no
&&\|x_{i}-x_{j}\|^2\\ \no
&&\;=\langle x_{i}-x_{j},x_{i}-x_{j}\rangle\\ \no
&&\;=\langle(\hat{z}_{i}+\tilde{z}_{i})-(\hat{z}_{j}+\tilde{z}_{j}),(\hat{z}_{i}
+\tilde{z}_{i})-(\hat{z}_{j}+\tilde{z}_{j})\rangle\\ \no
&&\;=\langle \hat{z}_{i}-\hat{z}_{j},\hat{z}_{i}-\hat{z}_{j}\rangle
+\langle\tilde{z}_{i}-\tilde{z}_{j},\tilde{z}_{i}-\tilde{z}_{j}\rangle\\ \no
&&\;=\|\hat{z}_{i}-\hat{z}_{j}\|^2+\|\tilde{z}_{i}-\tilde{z}_{j}\|^2\\ \label{eq:1}
&&\;\ge\|\hat{z}_{i}-\hat{z}_{j}\|^{2}
+\left|\|\tilde{z}_i\|-\|\tilde{z}_j\|\right|^2,
\en
where the Cauchy-Schwarz inequality is used to obtain the last inequality.
%
The following results follow easily from \eqref{eq:1}.

\begin{theorem}\label{t:1}
Given two points $x_{i},x_{j}\in D$ with the form (\ref{x-split}),

$(i)$ if $\|\hat{z}_i-\hat{z}_j\|>\vep$
then $x_i\notin N_{\vep}(x_j)$, where $N_{\vep}(x_j)$ is an $\vep$-neighborhood of $x_j$;

$(ii)$ if $\|\hat{z}_{i}-\hat{z}_{j}\|^2+\left|\|\tilde{z}_i\|-\|\tilde{z}_j\|\right|^2>\vep^2$
then $x_{i}\notin N_{\vep}(x_{j})$.
\end{theorem}

By Theorem \ref{t:1} (i), if $\|\hat{z}_i-\hat{z}_j\|>\vep$ then it is known that $x_i$ is not in the
$\vep$-neighborhood of $x_j$ without calculating the distance between $x_i$ and $x_j$.
The time complexity of calculating the distance $\|x_{i}-x_{j}\|$ for each pair of points $x_i,x_j\in D$
is $O(n^{2}h)$, whilst that of calculating $\|\hat{z}_i-\hat{z}_j\|$ is $O(n^{2}h_1)$ for each pair
of points $\hat{z_i},\hat{z_j}$ corresponding to the points $x_i,x_j\in D$.
Since $\hat{z_i}$ and $\hat{z_j}$ are the projections of $x_i$ and $x_j$ on the space spanned by the
eigenvectors $\phi_1,\ldots,\phi_{h_1}$ corresponding to the largest $h_1$ eigenvalues and $h_1$ is usually
much smaller than $h$, the calculation of $\|\hat{z_i}-\hat{z_j}\|$ is faster than that of
$\|x_{i}-x_{j}\|$. In fact, the value of $h_1$ can be determined to be the smallest number of $d$ such that
the following inequality is satisfied for a given real number $p$:
\be\label{eq:6}
\frac{\sum_{i=1}^{d}\lambda_{i}}{\sum_{j=1}^{h}\lambda_{j}}\ge p,
\en
where $\lambda_{i}$ is the $i$th largest eigenvalue.
In the experiments conducted in Section \ref{sec4}, $p$ can be taken as $0.7,0.8,0.9$ or $0.99$, and
in such cases, $h_1$ is smaller than $h/2$ for most of the data sets.
This means that Theorem \ref{t:1} (i) can be used to exclude the data points $x_i$'s that are not in
the $\vep$-neighborhood of $x_j$ with a lower computational cost by only verifying
whether or not $\|\hat{z}_i-\hat{z}_j\|>\vep$ for their projections $\hat{z}_i$ and $\hat{z}_j$ on
the space spanned by the eigenvectors $\phi_1,\ldots,\phi_{h_1}$ corresponding to the largest $h_1$
eigenvalues.

%
%

On the other hand, in the case when $\|\hat{z_i}-\hat{z_j}\|<\vep$, we do not know if $x_i$ is still
in the $\vep$-neighborhood of $x_j$. However, this can be justified with Theorem \ref{t:1} (ii)
by further calculating $\left|\|\tilde{z_i}\|-\|\tilde{z_j}\|\right|^2$ with a much lower cost.
In fact, since
\ben\label{eq:2}
\|\tilde{z}_{i}\|^2&=&\langle\tilde{z}_{i},\tilde{z}_{i}\rangle
=\langle x_{i}-\hat{z}_{i},x_{i}-\hat{z}_{i}\rangle\\
&=&\langle x_{i},x_{i}\rangle+\langle\hat{z}_{i},\hat{z}_{i}\rangle-2\langle x_{i},\hat{z}_{i}\rangle\\
&=&\langle x_{i},x_{i}\rangle-\langle\hat{z}_{i},\hat{z}_{i}\rangle\\
&=&\|x_{i}\|^{2}-\|\hat{z}_{i}\|^{2},
\enn
we have
\be\label{eq:3}
\|\tilde{z}_{i}\|=\left(\|x_{i}\|^{2}-\|\hat{z}_{i}\|^2\right)^{1/2}.
\en
This means that $|\|\tilde{z_i}\|-\|\tilde{z_j}\||$ can be calculated with $O(nh)$ time-complexity.
In fact, $|\|\tilde{z_i}\|-\|\tilde{z_j}\||$ can be calculated and stored in advance.

By the discussions above it is known that Theorem \ref{t:1} can be used to exclude the data points $x_i$'s
that are not in the $\vep$-neighborhood of $x_j$ with the time complexity $O(n^{2}h_1)+O(nh)$
which is lower than $O(n^{2}h)$ since, in general, $h_1<<h$. However, if the conditions in
Theorem \ref{t:1} (i) and (ii) are not satisfied, then we need to calculate the distance
between $x_i$ and $x_j$ to see if $x_i$ is in the $\vep$-neighborhood of $x_j$.

For a single-range query, Theorem \ref{t:1} helps to reduce the time complexity
in the neighborhood finding process from $O(n^2h)$ to $O(n^{2}h_1)+O(nh)$.
This is a good improvement when $h=O(n)$ and $h_1<<h$. However, if $n$ is large enough, the $O(n^2)$
time-complexity is still quite high, and so it is necessary to further reduce the time complexity
by reducing the number of accesses during range query, that is, we use Theorem \ref{t:1} to deal with
only small part of the other $n-1$ points in finding the $\vep$-neighborhood for each point
during a single-range query. This can be done by introducing a reference point, as discussed below.

Suppose $z_{p_i,1}$ is the projection of $x_{p_i}\in D$ in the first principal component $\phi_1$
obtained by Algorithm \ref{alg:fpca}, $i=1,\ldots,n$, and $z_{p_1,1},z_{p_2,1},\ldots,z_{p_n,1}$
are arranged in decreasing order.
If $|z_{p_i,1}-z_{p_j,1}|>\vep$ for some positive integers $i,j$ with $j\le i\le n$,
then we have $|z_{p_j,1}-z_{p_k,1}|>\vep$
for any integer $k$ with $i\leq k\leq n$ and so, by Theorem \ref{t:1} (i),
$x_{p_k}\notin N_{\vep}(x_{p_j})$ for $i\leq k\leq n$.
As a result, we do not need to consider the point $x_{p_k}$ for $j\leq k\leq n$ when we do
the range query for $x_{p_j}$, that is, when we search for the $\vep$-neighborhood of $x_{p_j}$.
Hence, the number of accesses in range query is reduced. Further, The calculation of the
distance between $x_{p_k}$and $x_{p_j}$ can be pruned in batch for all integers $k$'s with $j\leq k\leq n$.

\begin{figure}[ht]
\centering
\subfigure[]{\includegraphics[width=0.37\textwidth]{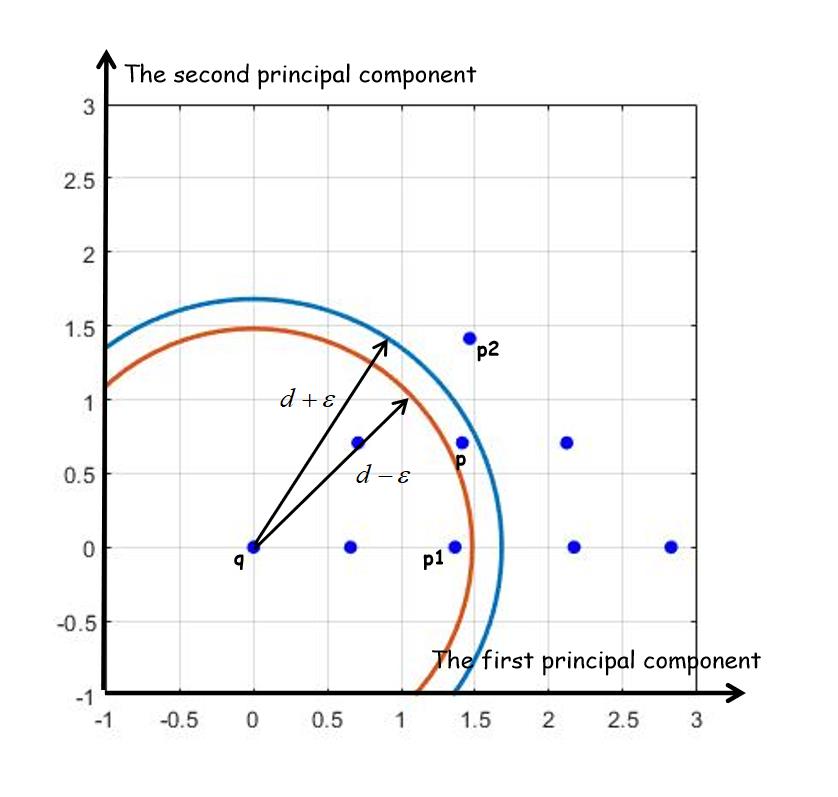}}
\subfigure[]{\includegraphics[width=0.37\textwidth]{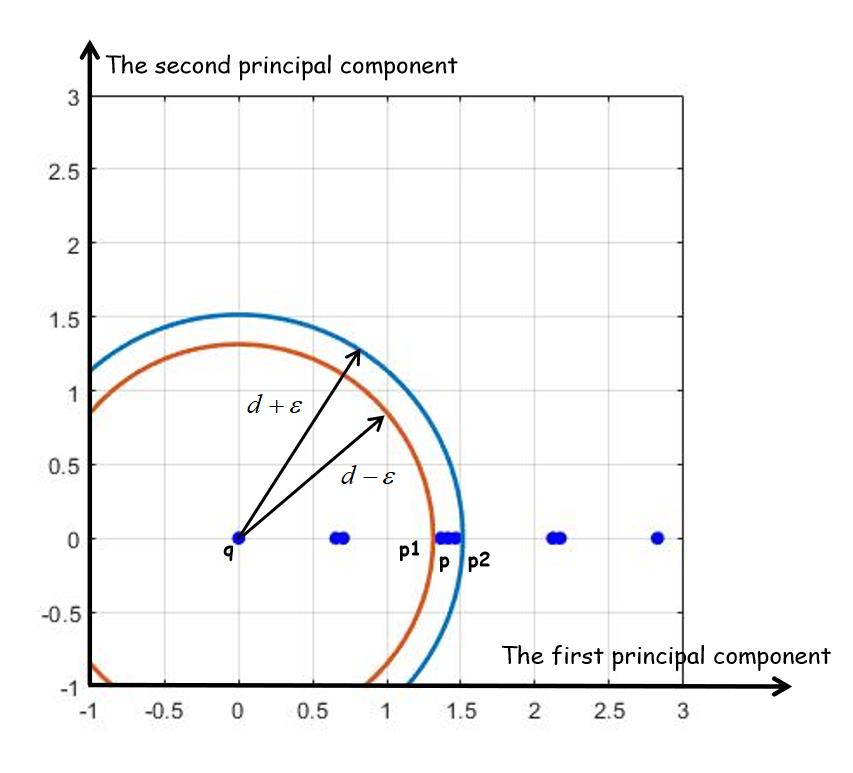}}
\caption{An example for the phenomenon of the "overcrowding effect" and the choice of reference points
}\label{fig:why2d}
\end{figure}

In the above process, only the projection on the first principal component $\phi_1$ is considered
for each data point. Unfortunately, this may lead to a wrong conclusion that the distance of the
projections $z_i,z_j$ on the first principal component $\phi_1$ of certain data points $x_i,x_j\in D$
is smaller than $\vep$, but the distance between the points $x_i,x_j$ may actually be quite large,
especially when the dimension $d$ of the data points is high.
A natural way to address this issue is to consider projections on the first $n_a$ principal components $\phi_1,\phi_2,\ldots,\phi_{n_a}$ for the data points with the integer $n_a\ge1$,
as illustrated in Figure \ref{fig:why2d}.
However, it is not easy to extend the process discussed in the preceding paragraph from the
projections on the first principal component to those on the first several principal components.
We will do this by introducing a reference point.
%

Note that Figure \ref{fig:why2d} (a) presents the projections on the first two principal components of
the data points represented by the blue points,
whilst Figure \ref{fig:why2d} (b) shows the projections on the first principal components of these
data points. From Figure \ref{fig:why2d} it is clear that the projections on the first two principal
components of the data points are very well separated (see Figure \ref{fig:why2d} (a)),
but the projections on the first principal component of some data points are very close or even almost
coincide (see Figure \ref{fig:why2d}(b)).
%

For an integer $n_a\ge1$ let $q=\sum_{j=1}^{n_a}q_{j}\phi_j$ be a reference point. Hereafter,
for convenience, we call such $q$ an $n_a$-dimensional point and write $q=(q_1,\ldots,q_{n_a})$.
Set
\be\label{eq:4.1}
\hat{z_{i}}=z_{i}^{0}+z_{i}^{1},
\en
where $z_i^0=\sum_{k=1}^{n_a}z_{i,k}\phi_k$ and $z_i^1=\sum_{k=n_a+1}^{h_1}z_{i,k}\phi_k$.
Then
\be\label{eq:4.2}
\|\hat{z_{i}}-\hat{z_{j}}\|^{2}=\|z_{i}^{0}-z_{j}^{0}\|^{2}+\|z_{i}^{1}-z_{j}^{1}\|^{2}
\en
with
\be\no
\|z_i^0-z_j^{0}\|&=&\|z_{i}^0-q+q-z_{j}^0\|\\ \label{eq:4.3}
&\ge&\left|\|z_i^{0}-q\|-\|z_{j}^{0}-q\|\right|=|d_i-d_j|,
\en
where
\be\label{eq:5}
d_i:=\|q-z_i^0\|,\;\;\;i=1,\ldots,n.
\en
%
%
%
%
%
Rearrange $d_1,\ldots,d_n$ in decreasing order as $d_{p_{1}},d_{p_{2}},\ldots,d_{p_{n}}$ and set $d=[d_{p_{1}},d_{p_{2}},\ldots,d_{p_{n}}]$. Then we have the following result
which can be used to prune unnecessary distance calculations for a batch of points at once.

\begin{theorem}\label{c:1}
For any two different points $x_{p_{m}},x_{p_{k}}\in D$ with $0<m<k\leq n$,
if $|d_{p_m}-d_{p_k}|>\vep$ then $\{x_{p_l}, n\ge l\ge k\}\not\subset N_{\vep}(x_{p_m})$
and $\{x_{p_l},0<l\le m\}\not\subset N_{\vep}(x_{p_k})$.
\end{theorem}

\begin{proof}
By \eqref{eq:1}, \eqref{eq:4.1}, \eqref{eq:4.2}, \eqref{eq:4.3} and \eqref{eq:5}
it follows that for $n\ge l\ge k$ we have
\ben
\|x_{p_l}-x_{p_m}\|&\ge&\|\hat{z}_{p_l}-\hat{z}_{p_m}\|\ge\|z_{p_l}^0-z_{p_m}^0\|\\
&\ge& |d_{p_l}-d_{p_m}|\ge|d_{p_k}-d_{p_m}|>\vep.
\enn
This implies that $x_{p_l}\notin N_{\vep}(x_{p_{m}})$.
The case $0<l\le m$ can be proved similarly.
\end{proof}

Figure \ref{fig:why2d} (a) illustrates the significance of choosing a two-dimensional reference
point in pruning unnecessary distance calculations.
Suppose we want to search for the $\vep$-neighborhood of the point whose projection on the first
two principal components $\phi_1,\phi_2$ is $p$.
The annular region associated with $p$ is given by
\ben
R_p=\{z=\sum_{j=1}^{2}z_j\phi_j\;:\;d-\vep\le\|z-q\|\le d+\vep\},
\enn
where $d=\|p-q\|$ and $q$ is a two-dimensional reference point defined as $q=\sum_{j=1}^{2}q_j\phi_j$.
By Theorem \ref{c:1} there is no need to access all the points whose projections on $\phi_1,\phi_2$
are outside of the annular region $R_p$ in the process of finding an $\vep$-neighborhood of $p$.

\begin{figure}
\centering
\includegraphics[width=0.45\textwidth]{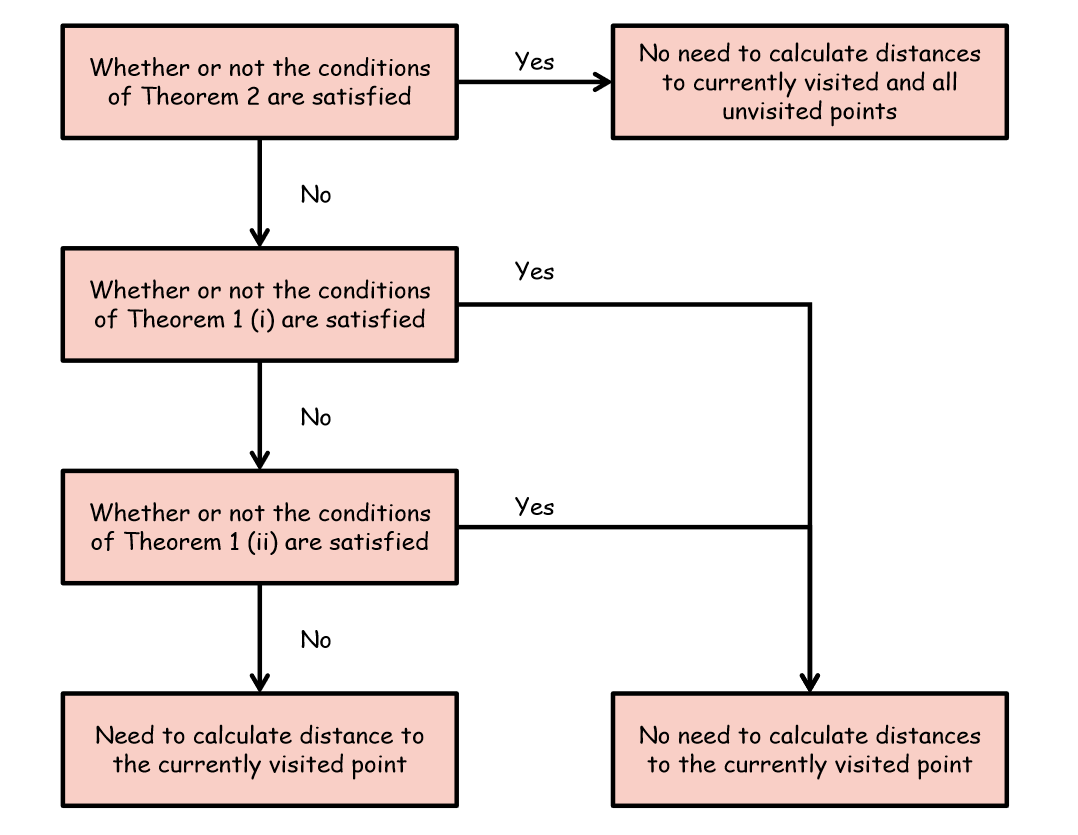}
\caption{The main idea of the neighborhood finding process.
}\label{fig:diagram2}
\end{figure}

Based on Theorems \ref{t:1} and \ref{c:1}, we propose a Fast PCA Pruning (FPCAP) algorithm to accelerate
the range query for $x_{p_m}\in D$ in neighborhood finding. The main idea of the FPCAP algorithm is shown
in Figure \ref{fig:diagram2}, and the detailed algorithm is given in Algorithm \ref{alg:rq}.
Algorithm \ref{alg:rq} can be roughly divided into four stages. Stage I excludes all the points on
one side of $x_{p_{m+t}}$, that is, Stage I reduces the redundant distance calculations in batches.
Stages II and III exclude the point at which the current iteration is reached
if the condition in Step 4 or Step 6 in Algorithm \ref{alg:rq} holds.
In Stage IV, we have to calculate the distance between $x_{p_{m}}$ and $x_{p_{m+t}}$ to see if
$x_{p_{m+t}}$ is in the neighborhood of $x_{p_m}$:
\ben
d_{p_{m},p_{m+t}}=\|x_{p_m}-x_{p_{m+t}}\|.
\enn
When we do range query for $x_{p_k}$, the parameter $\rm{step}=1$ or $-1$ in Algorithm \ref{alg:rq}
means finding neighbors of $x_{p_k}$ from the set $\{x_{p_j}|k<j\leq n\}$ or $\{x_{p_j}|0<j<k\}$.

\begin{algorithm}
\caption{FPCAP}\label{alg:rq}
\hspace*{0.02in}
{\bf Input:} row data matrix $X=[x_1,x_2,\ldots,x_n]^T$ of size $n\times h$,
projection data matrix $Z=[z_1,z_2,\ldots,z_n]^T$ of size $n\times h$,
candidate point $x_{p_{m}}$, reference point $q$, $\vep$, $\rm{step}$, $h_1$,
$t=0$, $h'=1$, $\rm{Diff}=0$. \\
\hspace*{0.02in}
{\bf Output:} $\vep$-neighborhood $N_{\vep}(x_{p_{m}})$ of $x_{p_{m}}$.\\
\vspace{-0.1in}
\begin{algorithmic}[1]
	\STATE Calculate $d_i=\|q-z_i^0\|$ for $i=1,\ldots,n$, and rearrange them in decreasing order as
           $d_{p_1},d_{p_2},\ldots,d_{p_n}$.
	\STATE $t=t+\rm{step}$
	\STATE If $m+t>n$, stop.
	\STATE Stage I: Calculate
	    \ben
		d=|d_{p_{m}}-d_{p_{m+t}}|.
		\enn
		If $d>\vep$, stop.
	\STATE Stage II: Calculate
		\ben
		\rm{Diff}:=\rm{Diff}+\|z_{p_{m},h'}-z_{p_{m-t},h'}\|^2.
		\enn
	    If $\rm{Diff}>\vep^{2}$, go to Step 2.
	\STATE Update $h'=h'+1$. If $h'\leq h_{1}$, go back to Step 5.
	\STATE Stage III: Calculate
		\ben
		&&\|\tilde{z}_{p_{m}}\|_2=(\|x_{p_{m}}\|^2-\|\hat{z}_{p_{m}}\|^2)^{1/2},\\
		&&\|\tilde{z}_{p_{m+t}}\|=(\|x_{p_{m+t}}\|^2-\|\hat{z}_{p_{m+t}}\|^2)^{1/2}.
		\enn
		If $\rm{Diff}+(\|\tilde{z}_{p_m}\|-\|\tilde{z}_{p_{m+t}}\|)^2>\vep^2$, go to Step 2.
	\STATE Stage IV: Calculate
		\ben
		d_{p_{m},p_{m+t}}=\|x_{p_{m}}-x_{p_{m+t}}\|.
		\enn
		If $d_{p_{m},p_{m+t}}\le\vep$, Update
		\ben
		N_{\vep}(x_{p_{m}})=N_{\vep}(x_{p_{m}})\cup \{x_{p_{m+t}}\}
		\enn
		and go to Step 2.
	\end{algorithmic}
\end{algorithm}

The reference point $q$ in Algorithm \ref{alg:rq} needs to be given in advance, and its choice
is essential for the effectiveness of the algorithm.
In this paper, the $n_{a}$-dimensional reference point $q$ ($n_a\ge1$) is chosen as
\be\label{refp}
q=(Z_{min,n_a},\ldots,Z_{min,n_a})=Z_{min,n_a}\sum_{j=1}^{n_a}\phi_j,
\en
where
\ben
 Z_{min,n_a}:=\min\{z_{i,j}:\;1\le i\le n,\;1\le j\le n_a\}
\enn
and $Z=(z_{i,j})_{n\times h}$ is the projection of the data obtained by Algorithm \ref{alg:fpca},
that is, $x_i=\sum_{j=1}^{h}z_{i,j}\phi_{j}$ for $x_i\in D$ with
$z_{i,j}=\langle x_i,\phi_j\rangle$, $i=1,\ldots,n,\;j=1,\ldots,h$
(see the beginning of this section). Thus
\be\no
d_i&=&\|q-z_i^0\|\\ \label{ref-d}
 &=&\left(\sum_{j=1}^{n_a}|z_{i,j}-Z_{min,n_a}|^2\right)^{1/2},\;i=1,\ldots,n.
\en

Note that the special case when $n_a=1$ is considered in \cite{lai2010fast} in a different context
where the Fast-PCA technique \cite{sharma2007fast} was applied to accelerate the global $k$-means
algorithm proposed in \cite{likas2003global}. In this case the one-dimensional reference point
$q=\min_{1\le i\le n}z_{i,1}$, where $z_{i,1}$ is the projection value of $x_i$ along the first
principal component $\phi_1$, that is, $z_{i,1}=\langle x_i,\phi_1\rangle$, so $d_i=|z_{i,1}-q|$,
$i=1,\ldots,n$.

\begin{figure}
\centering
\subfigure[]{\includegraphics[width=0.45\textwidth]{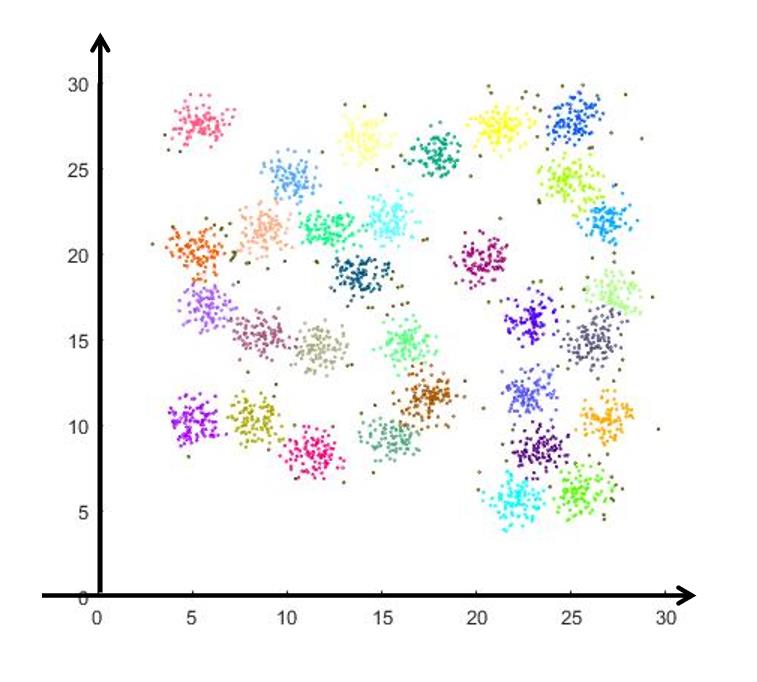}}
\subfigure[]{\includegraphics[width=0.45\textwidth]{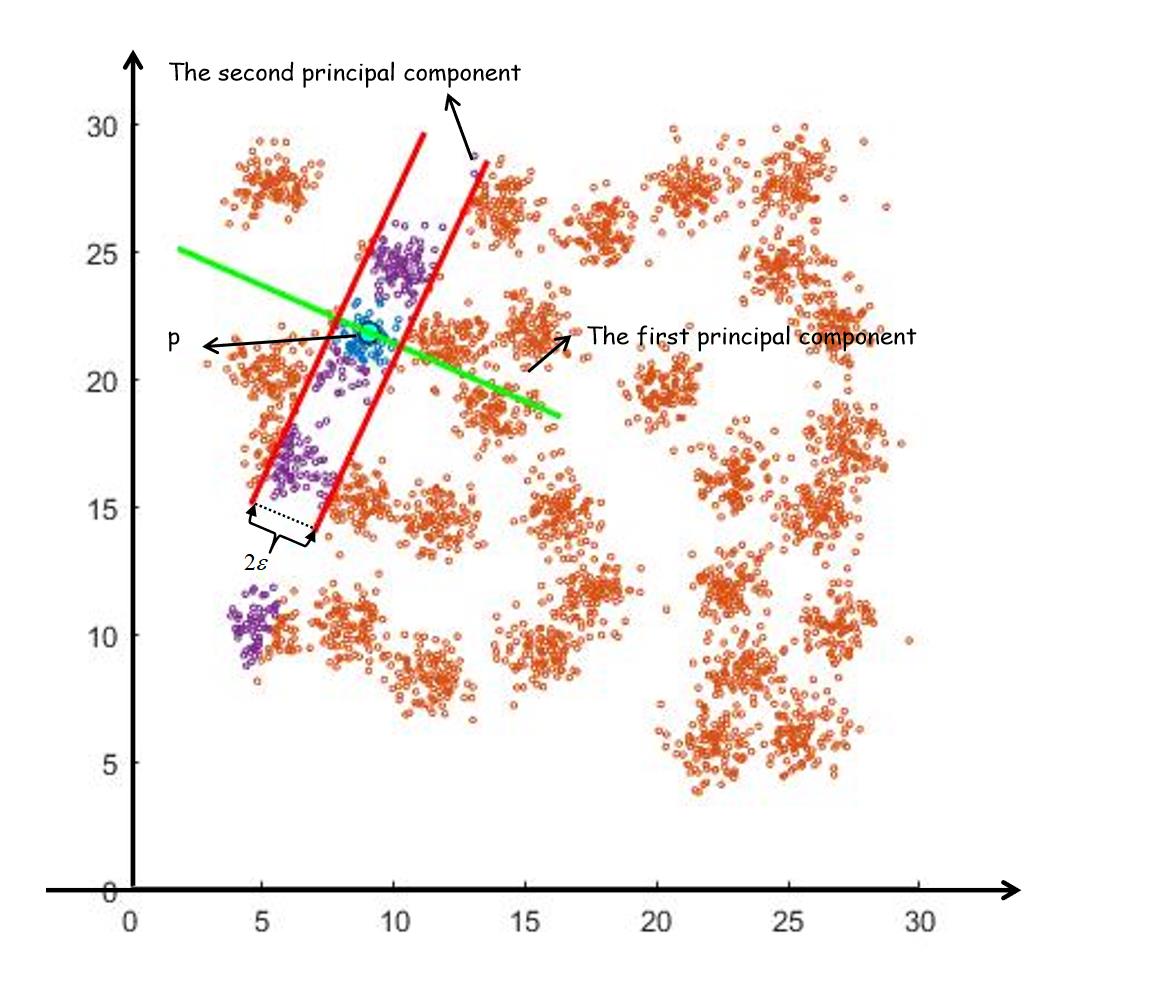}}
\caption{Illustration of pruning process in Stages I and II in Algorithm \ref{alg:rq}
}\label{fig:d31}
\end{figure}

Figure \ref{fig:d31} illustrates the pruning process in Stages I and II in Algorithm \ref{alg:rq}
on the D31 data set which is a two-dimensional synthetic data sets with $3100$ cardinalities and $31$ clusters.
Figure \ref{fig:d31} (a) shows the clustering results of the DBSCAN algorithm with $\vep=1.32$ and
$\textrm{MinPts}=68$. In Figure \ref{fig:d31} (b), the green and red lines represent the directions
of the first and second principal components, respectively.
The distances between the big green point $p$ and the two red lines both are $\vep$.
When we search for the $\vep$-neighborhood for $p$, the orange points and the purple points represent
the data points pruned in Stages I and II in Algorithm \ref{alg:rq}, while the blue points are the ones
whose distances to $p$ must be calculated. The results show that there are only a small number of points
whose distances to $p$ really need to be calculated. In other words,
Algorithm \ref{alg:rq} can prune unnecessary distance calculations effectively
when finding neighbors and estimating densities.

\subsection{An improved DBSCAN (IDBSCAN) algorithm}

\begin{figure*}
\centering
\includegraphics[width=0.95\textwidth]{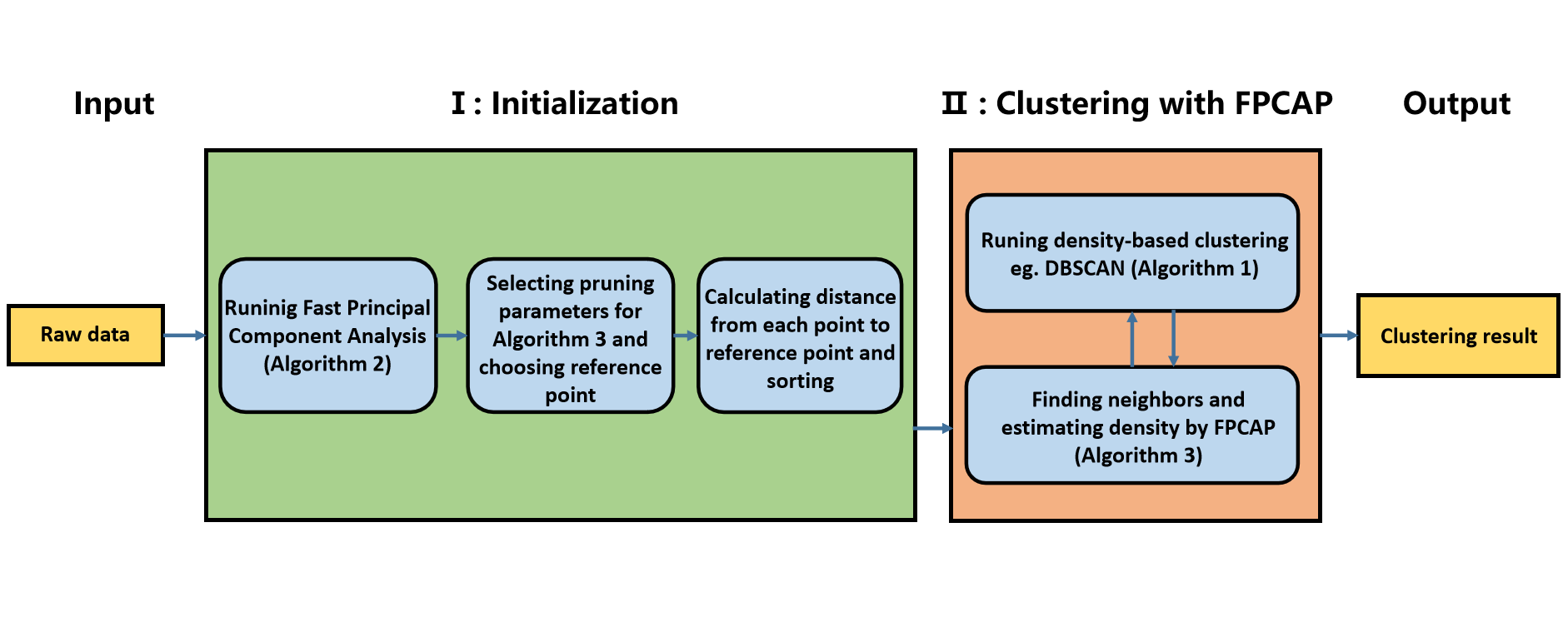}
\caption{The framework of the proposed FPCAP algorithm combined with a density-based clustering algorithm}
\label{fig:diagram}
\end{figure*}

The proposed FPCAP algorithm can be used together with a density-based clustering algorithm
to get an improved density-based clustering algorithm.
Figure \ref{fig:diagram} shows the framework of the proposed FPCAP algorithm combined with a
density-based clustering algorithm which consists of the initialization stage (Stage I) and
the clustering stage (Stage II).
The main part of the initialization stage is fast principal component analysis.
A new representation of the $h$-dimensional raw data $D=\{x_1,x_{2},\ldots,x_{n}\}$
can be obtained from this stage with $Z=(z_{p_i,j})_{n\times h}$ being the projection of data
obtained by Algorithm \ref{alg:fpca}.
The clustering stage is the combination of the density-based clustering algorithm and the FPCAP algorithm.
The final clustering results are obtained after this stage.

As an example of the above framework, in this subsection, the FPCAP algorithm is applied in the
neighborhood-finding process of the DBSCAN algorithm to get an improved DBSCAN (IDBSCAN) algorithm.
The detailed IDBSCAN algorithm is given in Algorithm \ref{alg:fDBSCAN}.

\begin{algorithm}
\caption{IDBSCAN}\label{alg:fDBSCAN}
\hspace*{0.02in}
{\bf Input:} row data matrix $X=[x_1,x_2,\ldots,x_n]^T$ of size $n\times h$, $\vep$, $\text{p}$\\
\hspace*{0.02in}
{\bf Output:} the clustering result
\begin{algorithmic}[1]
	\STATE Get the projection data matrix $Z=[z_1,z_2,\ldots,z_n]^T$ of size $n\times h$
           by Algorithm \ref{alg:fpca}
	\STATE Choose the reference point $q$ as in (\ref{refp})
	\STATE Run Algorithm \ref{alg:DBSCAN} (DBSCAN) with Algorithm \ref{alg:rq} (FPCAP)
           used in the neighborhood-finding process
	\STATE Return the clustering result
\end{algorithmic}
\end{algorithm}

\subsubsection{Correctness analysis}


In view of the indeterminacy in DBSCAN-like methods, border points may belong to different
clusters in the case when the order of the data points appeared in DBSCAN and IDBSCAN is different.
However, if the order of the data points appeared in DBSCAN and IDBSCAN is assumed to be the same,
then IDBSCAN and DBSCAN produce the same result.
On the other hand, by Theorems \ref{t:1} and \ref{c:1} (see also Figure \ref{fig:diagram2})
the conditions in Stages I, II and III in Algorithm \ref{alg:rq} are only used to justify
if $x_{p_{m+t}}\notin N_{\vep}(x_{p_{m}})$, and if the conditions in Stages I, II and III in
Algorithm \ref{alg:rq} are not satisfied then
we need to calculate the distance between the original points $x_{p_{m+t}}$ and $x_{p_{m}}$ to
see if $x_{p_{m+t}}\in N_{\vep}(x_{p_{m}})$ (see Stage IV in Algorithm \ref{alg:rq}).
Hence, IDBSCAN and DBSCAN obtain the same $\vep$-neighborhood result. Since the process of
IDBSCAN is the same as that of DBSCAN except for the different methods in finding $\vep$-neighborhoods,
IDBSCAN improves the efficiency of DBSCAN without losing correctness under the assumption that
the order of the data points appeared in DBSCAN and IDBSCAN is the same.
Even without the above assumption, DBSCAN and IDBSCAN still produce the same clustering result
if we only consider the clustering result of core points.

\subsubsection{Complexity analysis}

We focus on the time complexity analysis about the distance calculation which dominates the runtime of the range query in FPCAP (Algorithm \ref{alg:rq}). 
Denote by $n_1$ the total number of pruning in Stage I of Algorithm \ref{alg:rq}, and
denote by $n_2$ the total number of pruning for range query in Stages II and III of Algorithm \ref{alg:rq}.
Thus, on searching for the $\vep$-neighborhood for $x_{p_j}$ in Algorithm \ref{alg:rq},
if the condition $|d_{p_j}-d_{p_k}|>\vep$ is satisfied for $j<k\leq n$ then $n_1$ is increased
by $n-(k-1)$, and
if the conditions in Theorem \ref{t:1} are satisfied for $x_{p_j}$ and $x_{p_k}$ then $n_2$ is increased by $1$.
Define $n_{0}$ to be the average number of distance computations in Stage IV. Then
\be\label{n-split}
n_{0}:=(n-1)-\frac{n_{1}}{n}-\frac{n_{2}}{n}.
\en
The complexity of calculation of \eqref{eq:5} and \eqref{eq:3} is $O(nn_{a})$ and $O(nh)$, respectively.
The complexity of Stages II and IV is $O(n(n-{n_1}/{n})h_1)$ and $O(n(n-{n_1}/{n}-{n_2}/{n})(h-h_1))$,
respectively. As a result, the whole time complexity of FPCAP (Algorithm \ref{alg:rq}) is $O(nn_{a}+n(n-{n_1}/{n})h_1+n(n-{n_1}/{n}-{n_2}/{n})(h-h_1)+nh)$. In the case when $n-n_1/n=O(1)$,
the complexity is $O(nh)$, which is a very good improvement.
Take the experiments conducted on the subset of Reactionnetwork data set (n=20000) in
Section \ref{sec5} as example, when $n_{a}=1$ and $\vep=1000,10000,20000$ and $30000$,
${n_1}/{n}=19615,17125,14706$ and $12387$ and $n_{0}=110, 147, 452\ \text{and}\ 1162$.
When $\vep$ is small, the value of $n-{n_1}/{n}<<n$ and $n_0<<n$.
In addition, when $p=0.8$, $h1=h/7$ which is much smaller than $h$.
Note that the FPCAP (Algorithm \ref{alg:rq}) includes the sorting process of the distance from the data points to the reference point, and its complexity is $O(n\log n)$, but its runtime is very short compared to the overall runtime of FPCAP (Algorithm \ref{alg:rq}).

\section{Experimental Results}\label{sec4}

\subsection{Data sets}

We now conduct several experiments on four real-world data sets and three synthetic data sets to illustrate
the efficiency of the FPCAP and the IDBSCAN algorithms. Clickstream is a $12$-dimensional real-world
data set with $165474$ cardinalities which are composed of information on clickstream from online stores
offering clothing for pregnant women. Household is a $7$-dimensional real-world data set with $2049280$
cardinalities including all attributes except the temporal columns data and time.
Mocap is a $36$-dimensional real-world data set with $78095$ cardinalities, which has $5$ types of hand
postures from $12$ users. ReactionNetwork is a KEGG Metabolic Reaction Network (Undirected) Data set
of $65554$ data points of $28$ dimensions.
Dim-set consists of three synthetic datasets, Dim6, Dim10, Dim15, with Gaussian clusters and $4051, 6751$
and $10126$ cardinalities, respectively. The above real-world data sets and synthetic data sets can be
downloaded from the UCI Machine Learning repository\footnote{\url{http://archive.ics.uci.edu/ml/}} and
clustering datasets website\footnote{\url{http://cs.joensuu.fi/sipu/datasets/}}, respectively.
Duplicated data points and points with missing coordinates are deleted. Each attribute of the data
is normalized to $[0,100000]$.

\subsection{Algorithms}

The following algorithms are used in the experiments:
\begin{itemize}
\item FPCAP$_1$: Fast-PCA pruning initialized with the one-dimensional reference point $q$ given
                   by \eqref{refp} with $n_a=1$,
\item FPCAP$_2$: Fast-PCA pruning initialized with the two-dimensional reference point $q$ given
                   by \eqref{refp} with $n_a=2$,
\item KD tree: KD tree indexing technique on row data,
\item KD tree with PCA: KD tree indexing technique on projected data obtained with Fast-PCA
                       (Algorithm \ref{alg:fpca}).
\end{itemize}

All the experiments are conducted on a single PC with Intel Core 2.9GHz i7 CPU (16 Cores) and 32G RAM.

\subsection{Experimental results}

\subsubsection{Experiment 1: Effect of the choice of the reference point}

\begin{figure}
\centering
\subfigure[Dim6]{\includegraphics[width=0.22\textwidth]{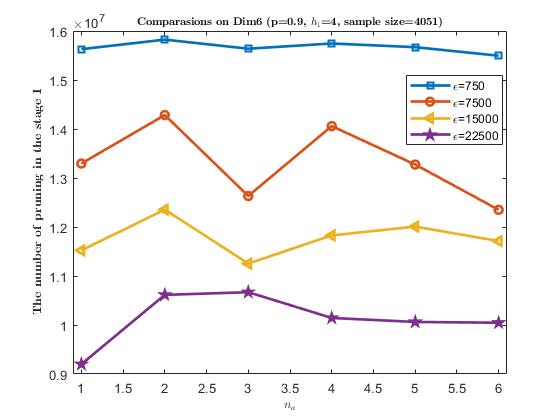}}
\subfigure[Dim10]{\includegraphics[width=0.22\textwidth]{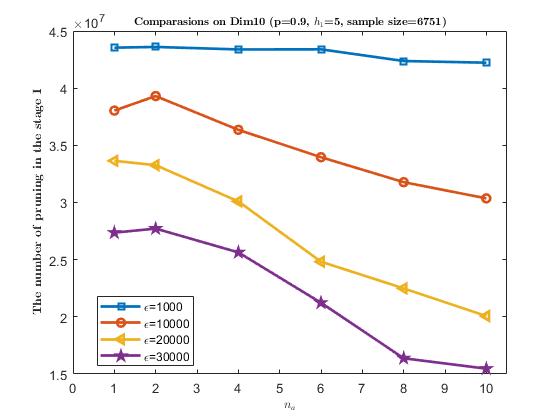}}
\subfigure[Dim15]{\includegraphics[width=0.22\textwidth]{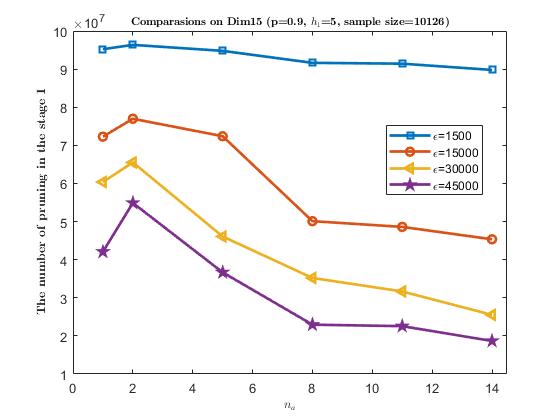}}
\subfigure[Clickstream]{\includegraphics[width=0.22\textwidth]{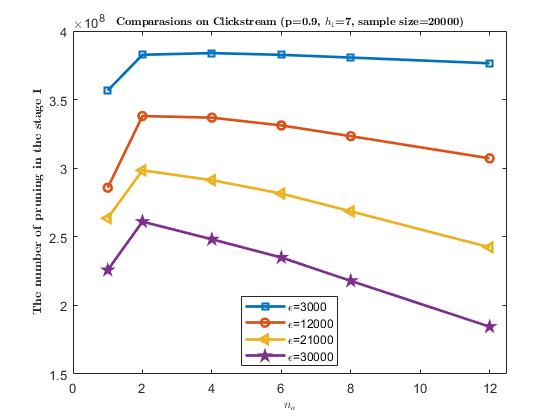}}
\subfigure[Household]{\includegraphics[width=0.22\textwidth]{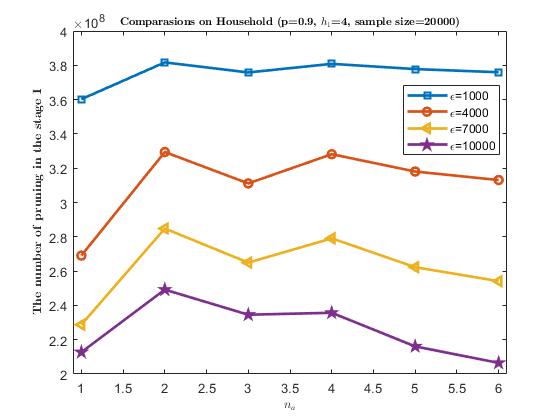}}
\subfigure[Mocap]{\includegraphics[width=0.22\textwidth]{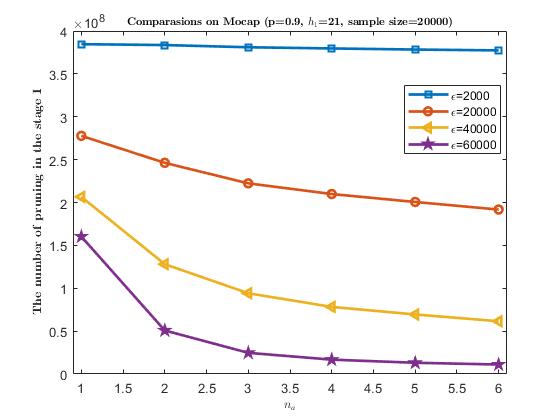}}
\subfigure[Reactionnetwork]{\includegraphics[width=0.22\textwidth]{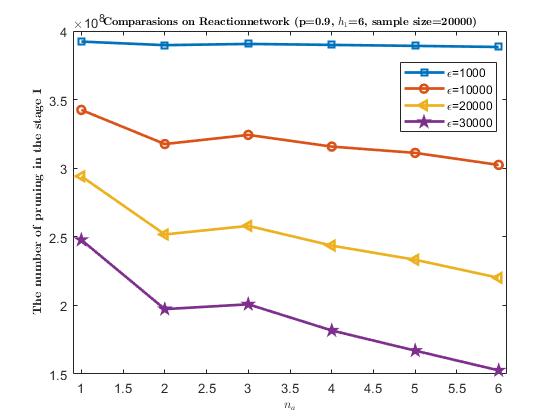}}
\caption{The number of pruning against the dimensionality $n_a$ of the reference point $q$
on seven data sets}
	\label{fig:distance_na}
\end{figure}

In the initialization process of FPCAP, the choice of the reference point is very important and
needs to be determined first. Experiment 1 is conducted to compare the effect on the pruning result
of the dimensionality $n_a$ of the reference point $q$.
The value of $n_1$ appearing in the expression \eqref{n-split}, which is the total number of pruning in
Stage I of Algorithm \ref{alg:rq}, determines the total number of accesses during range query,
that is, the total number of distance calculations in Stage II in Algorithm \ref{alg:rq} which dominates
the runtime of Algorithm \ref{alg:fDBSCAN}.
Therefore, Experiment 1 evaluates the initialization method in terms of the dimensionality
$n_a$ of the reference point $q$, based on $n_1$.
Figure \ref{fig:distance_na} presents the experimental results on the number
of pruning calculations in Stage I of Algorithm \ref{alg:rq} against the dimensionality $n_a$ of
the reference point $q$ for different neighborhood parameter $\vep$ on the three synthetic datasets,
Dim6, Dim10, Dim15, and the subsets of the four real-world data sets, Clickstream, Household, Mocap
and Reactionnetwork, with each subset containing $20000$ data points.
From Figure \ref{fig:distance_na} we have the following observations:
1) Initializing with a two-dimensional reference point (i.e., $n_a=2$) outperforms the other
cases with $n_a\not=2$ on most of the data sets used in the experiments such as Dim6, Dim10, Dim15,
Clickstream and Household for different choices of $\vep$; this means that the initialization with $n_a=2$
can effectively reduce the number of accesses during range query;
2) Initializing with a one-dimensional reference point (i.e., $n_a=1$) outperforms the other cases with
$n_a>1$ on Mocap and Reactionnetwork for different choice of $\vep$.
Based on the above observations, we use the cases $n_a=1,2$ to initialize
FPCAP (i.e., FPCAP$_1$ and FPCAP$_2$) in the remaining experiments.

\subsubsection{Experiment 2: Effect of the neighborhood radius $\vep$ in range query}

\begin{figure}
	\centering
	\subfigure[Dim6]{\includegraphics[width=0.22\textwidth]{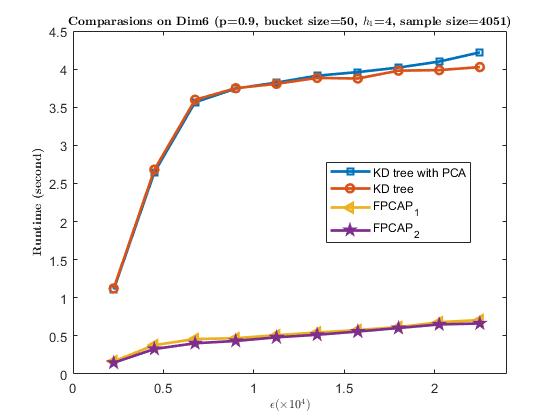}
	}
	\subfigure[Dim10]{\includegraphics[width=0.22\textwidth]{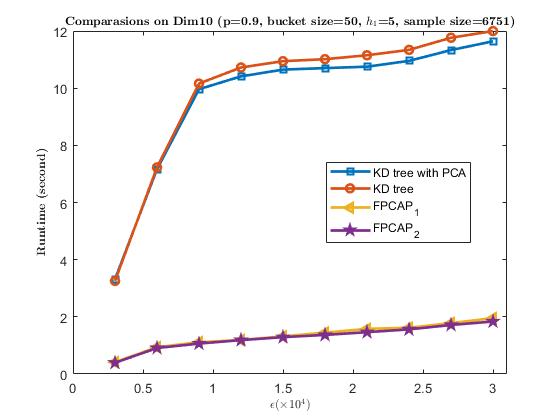}
	}
	\subfigure[Dim15]{\includegraphics[width=0.22\textwidth]{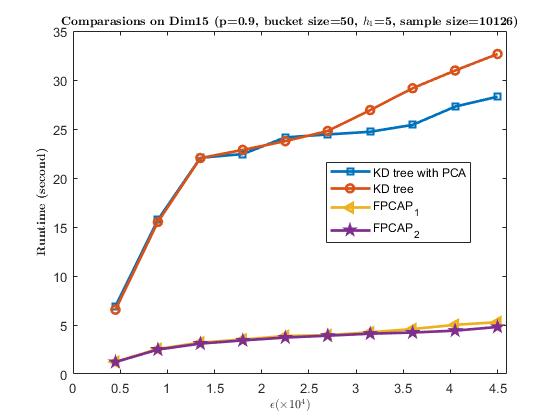}
	}
	\subfigure[Clickstream]{\includegraphics[width=0.22\textwidth]{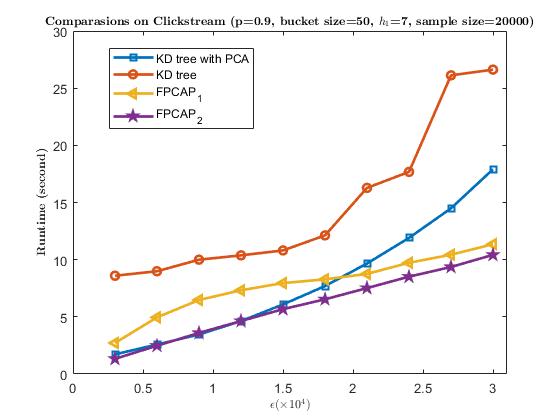}
	}
	\subfigure[Household]{\includegraphics[width=0.22\textwidth]{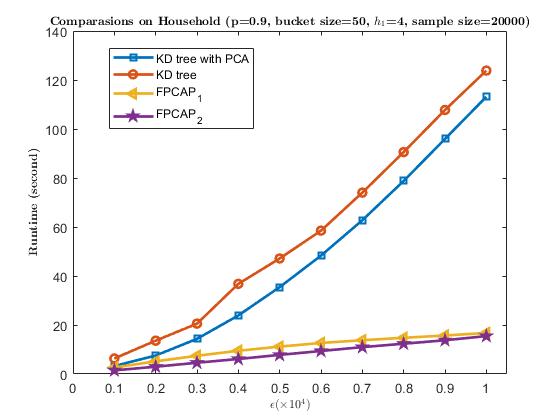}
	}
	\subfigure[Mocap]{\includegraphics[width=0.22\textwidth]{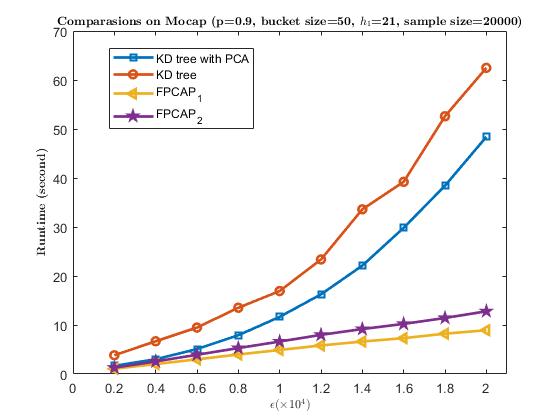}
	}
	\subfigure[Reactionnetwork]{\includegraphics[width=0.22\textwidth]{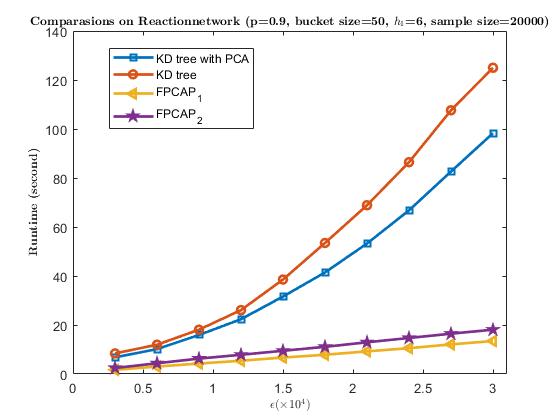}
	}
\caption{The runtime of the four algorithms on the seven data sets with different $\vep$
}\label{fig:runtime_eps}
\end{figure}

\begin{figure}
\centering
\subfigure[Dim6]{\includegraphics[width=0.22\textwidth]{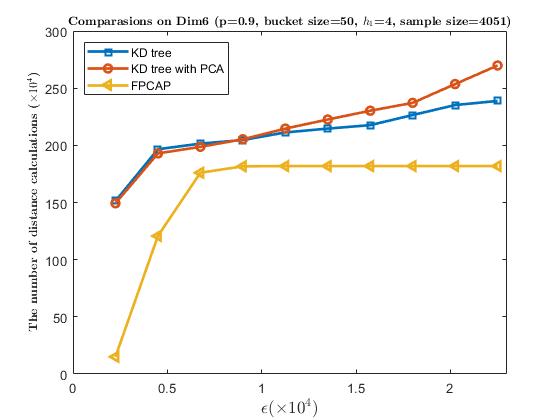}}
\subfigure[Dim10]{\includegraphics[width=0.22\textwidth]{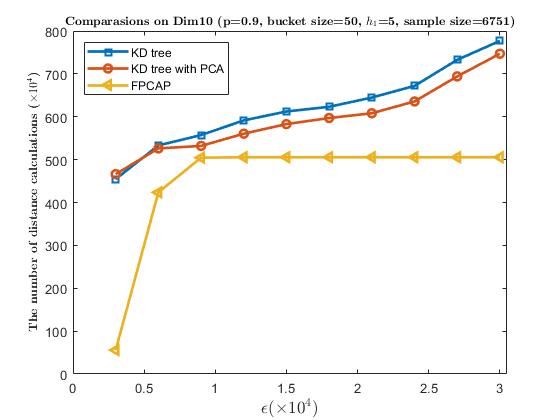}}
\subfigure[Dim15]{\includegraphics[width=0.22\textwidth]{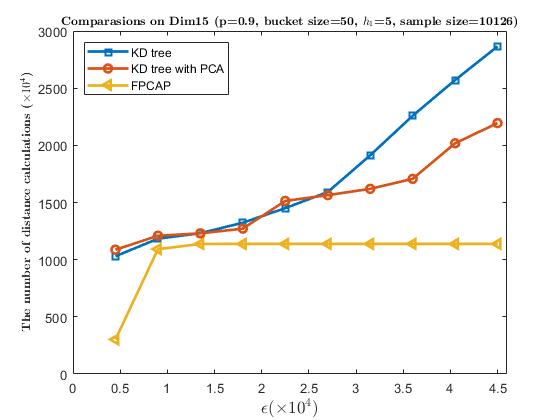}}
\subfigure[Clickstream]{\includegraphics[width=0.22\textwidth]{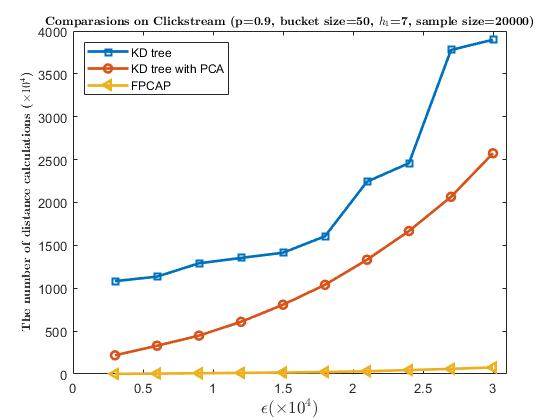}}
\subfigure[Household]{\includegraphics[width=0.22\textwidth]{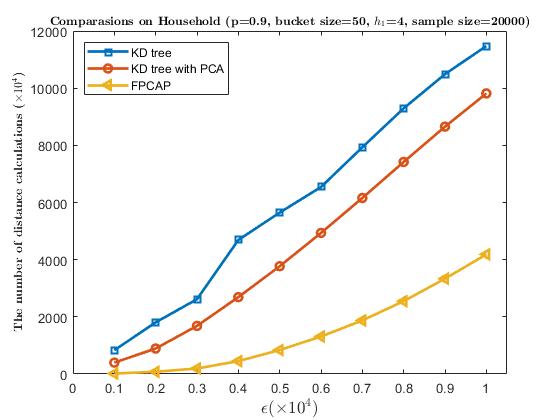}}
\subfigure[Mocap]{\includegraphics[width=0.22\textwidth]{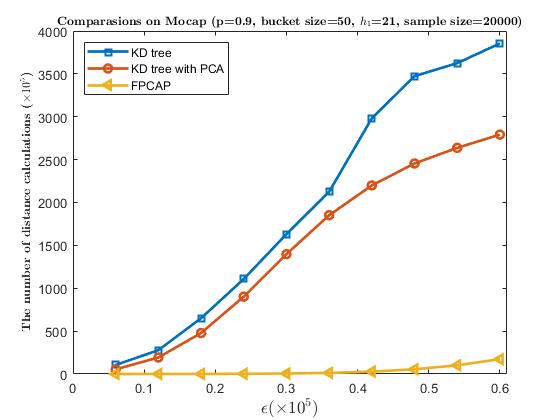}}
\subfigure[Reactionnetwork]{\includegraphics[width=0.22\textwidth]{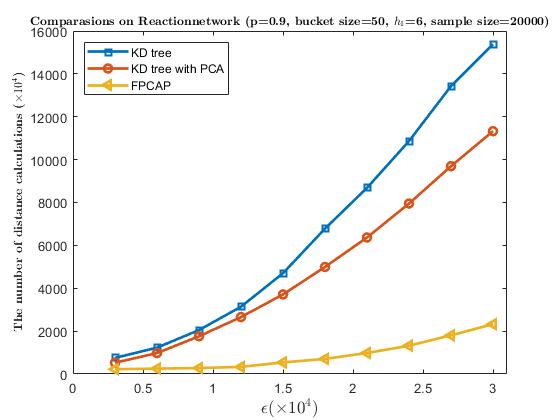}}
\caption{The number of distance calculations on the seven data sets with different $\vep$
}\label{fig:distance_eps}
\end{figure}

Experiment 2 compares the four algorithms, FPCAP$_1$, FPCAP$_2$, KD tree and KD tree with PCA,
on the same seven data sets as used in Experiment 1 for different $\vep$, in terms of their runtime
and $n_{0}$ (i.e., the total number of distance calculations).
Note that the runtime of the four algorithms was recorded as the average over $10$ duplicate tests to
reduce randomness.
Figure \ref{fig:runtime_eps} presents the runtime of the four algorithms against $\vep$.
The results show that FPCAP$_1$ and FPCAP$_2$ outperforms both KD tree and KD tree with PCA greatly
on the three synthetic data sets and three real-world data sets, Household, Mocap and Reactionnetwork,
and FPCAP$_1$, FPCAP$_2$ and KD tree with PCA outperforms KD tree on Clickstream with FPCAP$_2$
having the best performance among the four algorithms, KD tree with PCA performing better than FPCAP$_1$
when $\vep<2$ and FPCAP$_1$ having a better performance than KD tree with PCA when $\vep>2$.
Further, it is seen from Figure \ref{fig:runtime_eps} that the performance of FPCAP$_1$ and FPCAP$_2$
is similar on the seven data sets with FPCAP$_2$ performing slightly better than FPCAP$_1$ on the three
synthetic data sets and two real-world data sets, Clickstream and Household, and
FPCAP$_1$ performing slightly better than FPCAP$_2$ on Mocap and Reactionnetwork.
Furthermore, Figure \ref{fig:runtime_eps} shows that the runtime of KD tree and KD tree with PCA increases
much faster than FPCAP$_1$ and FPCAP$_2$ with $\vep$ increasing.
Figure \ref{fig:distance_eps} presents the value of $n_0$ (i.e., the total number of distance calculations)
of the pruning algorithms, KD tree, KD tree with PCA and FPCAP, against $\vep$.
The results in Figure \ref{fig:distance_eps} illustrate that, compared to the KD indexing technique,
FPCAP reduces more distance calculations.

\subsubsection{Experiment 3: Effect of the cardinality (data sample size) on the runtime}

\begin{figure}
\subfigure[Clickstream]{\includegraphics[width=0.22\textwidth]{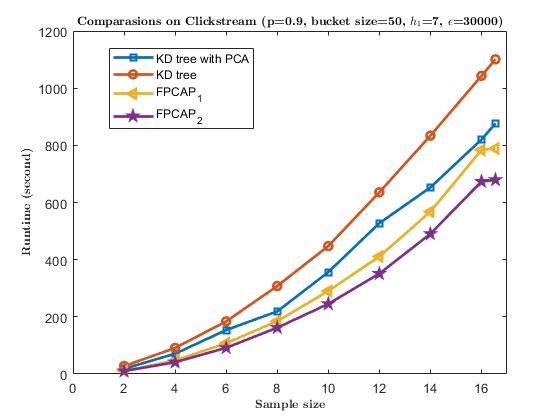}
}
\subfigure[Household]{\includegraphics[width=0.22\textwidth]{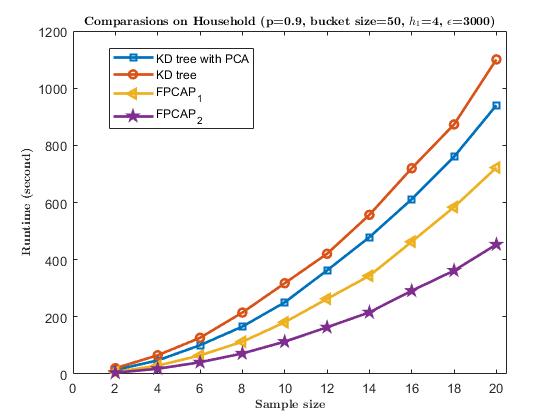}
}
\subfigure[Mocap]{\includegraphics[width=0.22\textwidth]{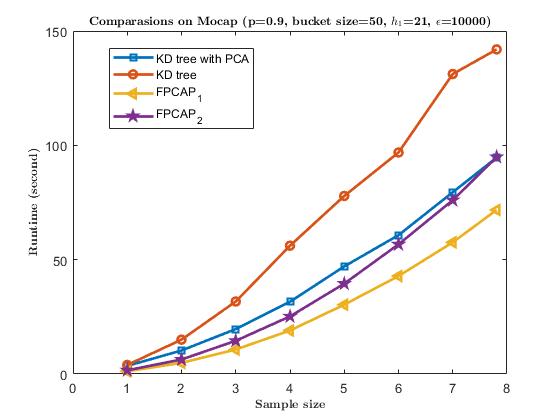}
}
\subfigure[Reactionnetwork]{\includegraphics[width=0.22\textwidth]{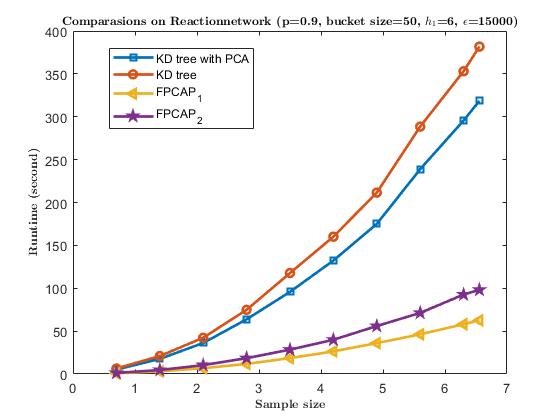}
}
\caption{
The runtime of the four algorithms against the cardinality (or data sample size) on the four
real-world data sets. The unit of the sample size is ten thousand.
}\label{runtime_n}
\end{figure}

Experiment 3 was conducted to compare the runtime of the four algorithms, FPCAP$_1$, FPCAP$_2$,
KD tree and KD tree with PCA, on the subsets of the four real-world data sets, Clickstream,
Household, Mocap and Reactionnetwork, for different cardinality (or data sample size).

Figure \ref{runtime_n} presents the runtime of FPCAP$_1$, FPCAP$_2$, KD tree and KD tree with PCA
against the data sample size $n$ on the subsets of the four real-world data sets.
The results in Figure \ref{runtime_n} show that FPCAP (FPCAP$_1$ and FPCAP$_2$) significantly
outperforms the KD tree indexing techniques (KD tree and KD tree with PCA)
as $n$ increases. It is further seen from Figure \ref{runtime_n} that the runtime of
KD tree and KD tree with PCA increases much faster than that of FPCAP$_1$ and FPCAP$_2$
with $n$ increasing.

\subsubsection{Experiment 4: Performance of FPCAP in combination with DBSCAN}

In this experiment we examine the performance of the Fast-PCA pruning algorithm, FPCAP, in combination
with DBSCAN. This we do by comparing with the naive way of neighbourhood finding and the KD tree indexing
technique in combination with the DBSCAN algorithm.
The five comparing algorithms are as follows:
\begin{itemize}
\item DBSCAN: The original DBSCAN algorithm (Algorithm \ref{alg:DBSCAN}) without pruning in the neighbourhood
finding process;
\item DBSCAN$_1$: DBSCAN (Algorithm \ref{alg:DBSCAN}) with the KD tree indexing technique on
the projected data obtained through Fast-PCA (Algorithm \ref{alg:fpca});
\item DBSCAN$_2$: DBSCAN (Algorithm \ref{alg:DBSCAN}) with KD tree indexing technique on row data;
\item IDBSCAN$_1$: IDBSCAN (Algorithm \ref{alg:fDBSCAN}) with $n_a=1$;
\item IDBSCAN$_2$: IDBSCAN (Algorithm \ref{alg:fDBSCAN}) with $n_a=2$.
\end{itemize}

\begin{figure}[h]
\centering
\includegraphics[width=0.5\textwidth]{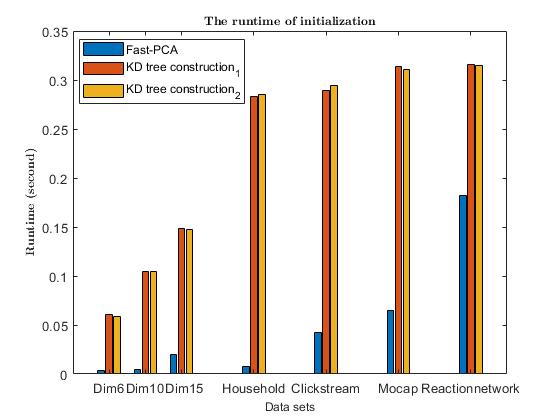}
\caption{The total runtime of performing Fast-PCA and KD tree construction, where KD tree construction$_1$
stands for the construction of KD tree on row data and KD tree construction$_2$ represents the construction
of KD tree on projected data obtained by Fast-PCA (Algorithm \ref{alg:fpca}).
}\label{fig:initialization}
\end{figure}

We first compare the runtime of the five comparing algorithms for different $\vep$, $\textrm{Minpts}$ and $p$.
Table \ref{tab:p} shows their runtime against $p$.
From Table \ref{tab:p} it is seen that (1) IDBSCAN has the best performance on all seven
data sets in terms of runtime;
(2) the KD tree indexing technique combined with Fast-PCA (DBSCAN$_1$) outperforms the original KD tree
indexing technique (DBSCAN$_2$) on all the four real-world data sets (except for the three synthetic
datasets, Dim6, Dim10, Dim15, on which the performance of the two algorithms are similar);
(3) for large $\vep$, DBSCAN with the KD tree indexing techniques (DBSCAN$_1$ and DBSCAN$_2$) need a much
longer runtime than the original DBSCAN algorithm does on the three synthetic data sets, Dim6, Dim10 and Dim15;
(4) the parameter $p$ varies from $0.7$ to $0.99$, but the runtime of IDBSCAN$_1$ and IDBSCAN$_2$ does
not change largely, meaning that the runtime of IDBSCAN$_1$ and IDBSCAN$_2$ is not
sensitive to the choice of the parameter $p$.
In addition, Figure \ref{fig:initialization}, which presents the total runtime of performing Fast-PCA and
KD tree construction, illustrates that the time used in initialization, that is, the runtime of Fast-PCA is
negligible compared with that of IDBSCAN and that the runtime of fast-PCA is much smaller than that of
the construction of the KD tree.

\begin{table*}
\centering
\begin{tabular}{cccccccc}
\toprule
Data sets& [MinPts, $\vep$] & DBSCAN  & DBSCAN$_1$ & DBSCAN$_2$ & $\text{p}\ (h_1)$
& IDBSCAN$_1$ & IDBSCAN$_2$ \\
\midrule
\multirow{6}{*} {Clickstream} &\multirow{3}{*}{[10, 15000]} & \multirow{3}{*}{22.09} & \multirow{3}{*}{6.45} & \multirow{3}{*}{11.33}
& 80\% (6)& 7.77&\textbf{5.83} \\
 & & & &  &90\% (7)& 8.13& \textbf{5.97} \\
 & & & &  &99\% (9)& 7.82& \textbf{5.96} \\
\cmidrule{2-8}
&\multirow{3}{*}{[20, 30000]} & \multirow{3}{*}{22.50} &\multirow{3}{*}{18.33} &\multirow{3}{*}{27.31}
& 80\% (6)& 11.72& \textbf{11.18} \\
 & & & &  &90\% (7)& 11.74&\textbf{11.13} \\
 & & & &  &99\% (9)& 11.67&\textbf{11.11} \\
\cmidrule{1-8}
\multirow{6}{*} {Household} &\multirow{3}{*}{[5, 1000]} & \multirow{3}{*}{19.88} & \multirow{3}{*}{3.47} &\multirow{3}{*}{6.66} &80\% (2)& 2.73& \textbf{1.59} \\
 & & & &  &90\% (4)& 2.76&\textbf{1.61} \\
 & & & &  &99\% (6)& 2.80&\textbf{1.63} \\
\cmidrule{2-8}
&\multirow{3}{*}{[10, 3000]} & \multirow{3}{*}{19.98} & \multirow{3}{*}{14.45}& \multirow{3}{*}{20.60}
&  80\% (2)& 7.59&\textbf{4.72} \\
 & & & &  &90\% (4)& 7.51&\textbf{4.89} \\
 & & & &  &99\% (6)& 7.48& \textbf{4.89} \\
\cmidrule{1-8}
\multirow{6}{*} {Dim6}  &\multirow{3}{*}{[3, 1500]} & \multirow{3}{*}{0.82} &\multirow{3}{*}{0.78} &\multirow{3}{*}{0.80} & 70\% (3)& 0.14& \textbf{0.13} \\
 & & & &  &90\% (4)& \textbf{0.13}& \textbf{0.13}\\
 & & & &  &99\% (6)& 0.15& \textbf{0.13}\\
\cmidrule{2-8}
&\multirow{3}{*}{[5, 5000]} & \multirow{3}{*}{1.04} & \multirow{3}{*}{3.15}& \multirow{3}{*}{3.15}
&  70\% (3)& 0.48& \textbf{0.43}\\
		& & & & &90\% (4)& 0.48& \textbf{0.44} \\
		& & & & &99\% (6)& 0.47& \textbf{0.41} \\
\cmidrule{1-8}
\multirow{6}{*} {Dim10}  & \multirow{3}{*}{[3, 2000]} & \multirow{3}{*}{2.52} &\multirow{3}{*}{2.62} &\multirow{3}{*}{2.48} &  80\% (3)& \textbf{0.34}& 0.36\\
		& & &  & &90\% (5)& \textbf{0.34}&  0.35\\
		& & &  & &99\% (8)& \textbf{0.34}&  0.35\\
\cmidrule{2-8}
&\multirow{3}{*}{[5, 6000]} & \multirow{3}{*}{2.89} & \multirow{3}{*}{7.43}& \multirow{3}{*}{7.45}
& 80\% (3)& 0.98&  \textbf{0.97}\\
		& & &  & &90\% (5)& 1.00&\textbf{0.96}\\
		& & &  & &99\% (8)& 1.01&\textbf{0.97}\\
\cmidrule{1-8}	
\multirow{6}{*} {Dim15}  & \multirow{3}{*}{[3, 3000]} & \multirow{3}{*}{6.34} &\multirow{3}{*}{5.65} &\multirow{3}{*}{5.28} &  80\% (4)& 1.01& \textbf{0.99}\\
		& & &  & &90\% (5)& 1.03& \textbf{0.96} \\
		& & &  & &99\% (8)& 1.03& \textbf{0.96} \\
\cmidrule{2-8}
&\multirow{3}{*}{[5, 9000]} & \multirow{3}{*}{7.14} & \multirow{3}{*}{16.34}& \multirow{3}{*}{16.11}
& 80\% (4)& 2.77&\textbf{2.63}  \\
		& & &  & &90\% (5)& 2.75& \textbf{2.63}\\
		& & &  & &99\% (8)& 2.69& \textbf{2.61}\\
\cmidrule{1-8}
\multirow{6}{*} {Mocap}  & \multirow{3}{*}{[5, 6000]} & \multirow{3}{*}{35.40} &\multirow{3}{*}{5.56} &\multirow{3}{*}{9.92} &  80\% (15)& \textbf{3.19}& 4.14\\
		& & &  & &90\% (21)& \textbf{3.30}& 4.07 \\
		& & &  & &99\% (31)& \textbf{3.29}& 4.17\\
\cmidrule{2-8}
&\multirow{3}{*}{[10, 14000]} & \multirow{3}{*}{35.85} & \multirow{3}{*}{23.18}& \multirow{3}{*}{34.92}
& 80\% (15)& \textbf{7.03}&9.21  \\
		& & &  & &90\% (21)& \textbf{6.84}& 9.18\\
		& & &  & &99\% (31)& \textbf{6.98}&9.22\\
\cmidrule{1-8}
\multirow{6}{*} {Reactionnetwork} & \multirow{3}{*}{[3, 5000]} & \multirow{3}{*}{31.40} &\multirow{3}{*}{9.37} &\multirow{3}{*}{10.95} &  80\% (4)& \textbf{2.79}& 3.97\\
		& & &  & &90\% (6)& \textbf{2.80}& 3.93 \\
		& & &  & &99\% (12)& \textbf{2.75}& 3.97 \\
\cmidrule{2-8}
&\multirow{3}{*}{[5, 10000]} & \multirow{3}{*}{30.94} & \multirow{3}{*}{18.55}& \multirow{3}{*}{21.08}
& 80\% (4)& \textbf{4.92}&7.30  \\
		& & &  & &90\% (6)& \textbf{4.89}& 7.08\\
		& & &  & &99\% (12)& \textbf{4.96}& 7.12\\
\bottomrule
\vspace{0.1cm}
\end{tabular}
\caption{Runtime (in seconds) of the five comparing algorithms on five benchmark data sets with different $p$.
}\label{tab:p}
\end{table*}

\subsection{Comprehensive analysis: Range query in high dimensions}

\begin{figure}
	\centering
	\subfigure[Case 1]{\includegraphics[width=0.45\textwidth]{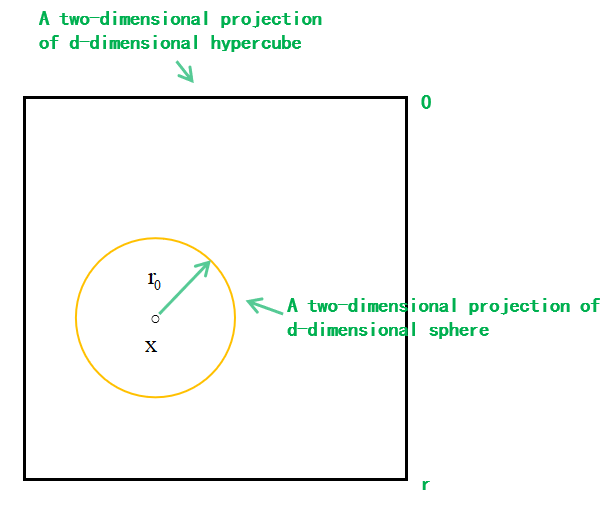}}
	\subfigure[Case 2]{\includegraphics[width=0.45\textwidth]{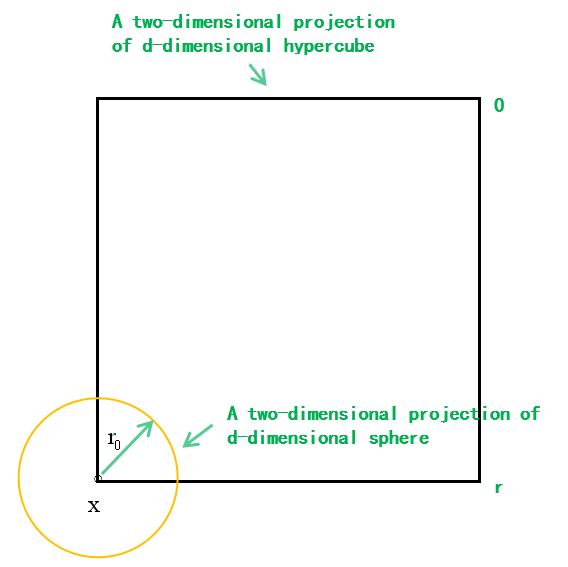}}
	\caption{Range query in high dimensions}
	\label{fig:case}
\end{figure}


\begin{theorem}\label{t:0}
Suppose the sample points $\{x_1,\ldots, x_n\}$ are uniformly distributed in a $d$-dimensional
hypercube $S$ with side length $r$. Denote by $R_{d,e}(x)$ the fraction of the sample points
captured by a $d$-dimensional sphere of radius $r_{0}$ and centered at $x\in S$, where $e=r_0/r$.
If $0\leq e\leq {1/2}$ then it follows that for any $x\in S$,
\ben
L_{d,e}\le R_{d,e}(x)\leq U_{d,e},
\enn
where $L_{d,e}=V_d(1)e^{d}2^{-d}$, $U_{d,e}=V_d(1)e^d$ and
$V_{d}(r)={\pi^{d/2}r^d}/\Gamma((d/2)+1)$ is the volume of a sphere of radius $r$
with the Gamma function $\Gamma(t)$ for $t>0$ satisfying that $\Gamma(t+1)=t\Gamma(t)$ 
and $\Gamma(1/2)=\sqrt{\pi}$.
\end{theorem}

\begin{proof}
Denote by $W_d(r)$ the volume of the hypercube with side length $r$. In the case when the sphere
of radius $r_0$ and centered at $x\in S$ is inside of the hypercube with side length $r$ without 
intersection, as seen in Figure \ref{fig:case} (a), we have $R_{d,e}(x)=V_d(r_0)/W_d(r)=V_d(1)e^d$
which is independent of $x\in S$ and thus is a upper bound of $R_{d,e}(x)$ for all $x\in S$.
In the case when the center $x\in S$ of the sphere locates at one of the vertices of the hypercube, 
as seen in Figure \ref{fig:case} (b), we have $R_{d,e}(x)=V_d(r_0)/[W_d(r)2^d]=V_d(1)e^d2^{-d}$
which is a lower bound of $R_{d,e}(x)$. 
\end{proof}

\begin{figure}
\centering
\includegraphics[width=0.5\textwidth]{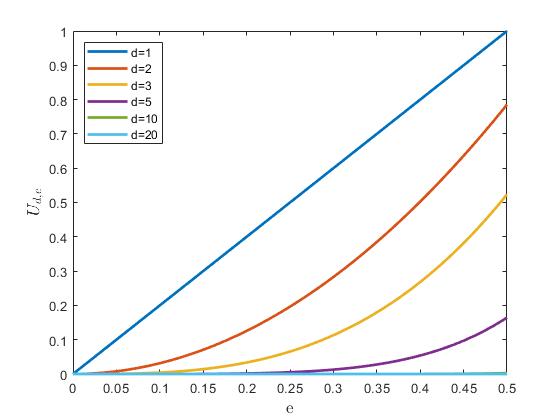}
\caption{The upper bound $U_{d,e}$ against $e$ for dimension $d=1,2,3,5,10,20$, where $U_{d,e}$ and $e$ 
are defined in Theorem \ref{t:0}. Note that 
when $e=1/2$, $U_{d,e}=1,0.7854,0.5236,0.1645,0.00249,2.46\times 10^{-8}$, corresponding to 
$d=1,2,3,5,10,20$, respectively.
}\label{fig:highdimension}
\end{figure}

Figure \ref{fig:highdimension} presents the relationship between the radius of the sphere and the upper bound
$U_{d,e}$ of $R_{d,e}(x)$ given in Theorem \ref{t:0}. The results illustrate the sparsity of high-dimensional 
data in the sense that 
for high-dimensional data which are uniformly distributed in a $d$-dimensional hypercube $S$ of side length $r$,
the number of the data points contained in a sphere of radius $r_0$ and centered at $x\in S$ is getting much 
smaller with $d$ increasing when $r_0/r\le1/2$.
For example, for sample points uniformly distributed in a $d$-dimensional hypercube $S$ 
of side length $r$, if $d>20$ then the sphere of radius $r/2$ and centered at $x\in S$ only contains 
$2.46\times 10^{-8}$ percent of the total number of sample points. Consider the process of the DBSCAN
algorithm. Due to the sparsity of high-dimensional data, the search radius of range query increases with
the dimension $d$. For the KD tree indexing technique, the larger the search radius is, the more sample points
need to be accessed in the process of range query, thus reducing the efficiency of range query.
Therefore, for the range query method combined with the KD tree indexing technique, when the search
radius is bigger than $r/2$, at least $50$ percent of the sample points need to be accessed in each range query.

\section{Conclusion}\label{sec5}

In order to accelerate the process of range query in density-based methods which are one of
the most popular clustering methods and have wide applications, 
we proposed a fast range query algorithm (called FPCAP), based on the 
fast principal component analysis which prunes unnecessary distance calculations in the range search 
process. By combining FPCAP with DBSCAN, we obtained an Improved DBSCAN (called IDBSCAN) algorithm. 
Experimental results on real-world and synthetic data sets demonstrate that both FPCAP
and IDBSCAN improve the computational efficiency and outperform other compared methods.
FPCAP can also be combined with other density-based clustering methods to improving their efficiency.


%

%

\ifCLASSOPTIONcaptionsoff
  \newpage
\fi

\bibliographystyle{IEEEtran}

\bibliography{IEEEabrv,refff}

\begin{IEEEbiography}[{\includegraphics[width=1in,height=1.25in,clip,keepaspectratio]{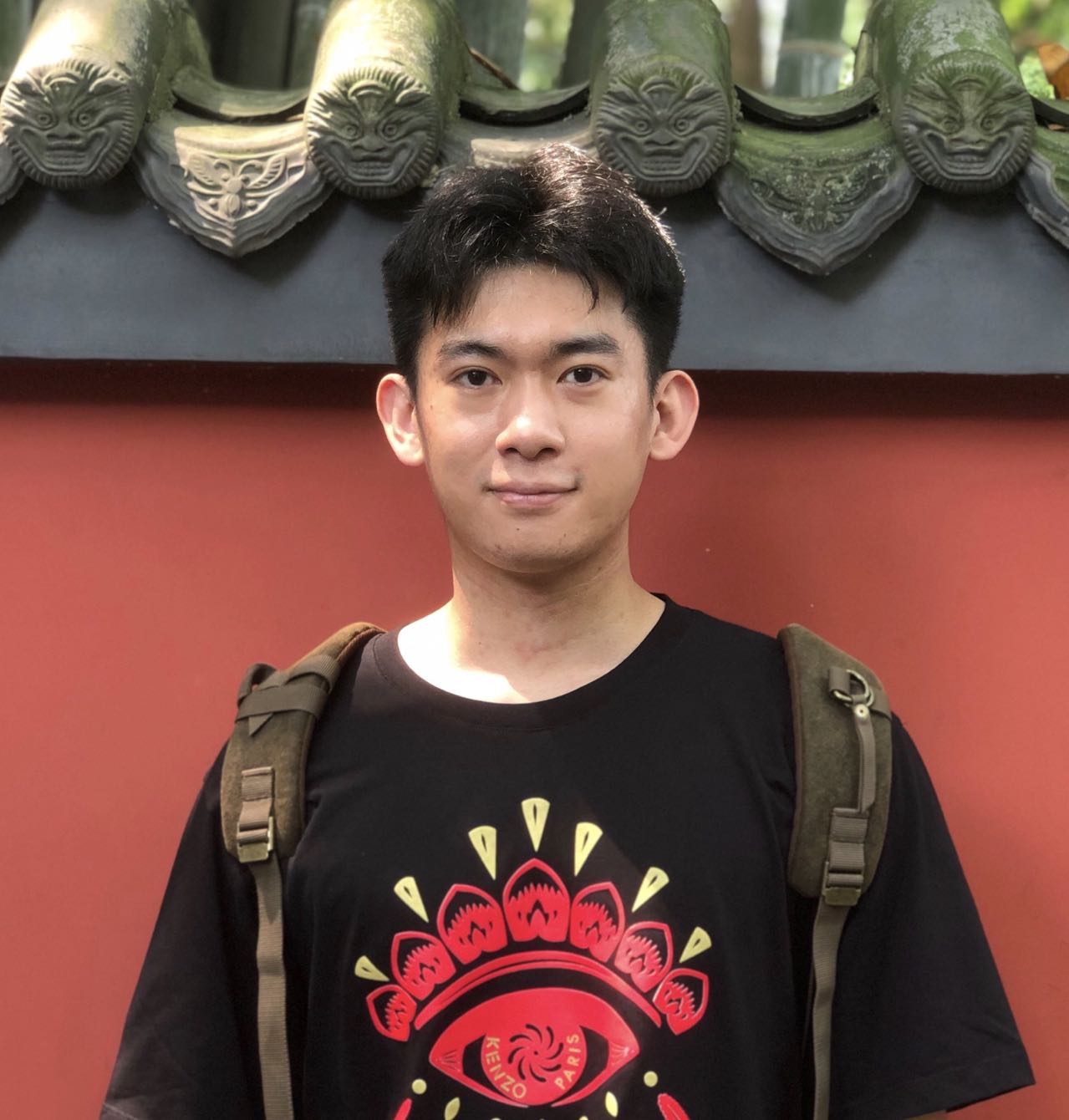}}]
{Difei~Cheng} received BSc degree in mathematics and applied mathematics from Shandong University,
Jinan, China, in 2017. He is currently pursuing his PhD degree in machine learning and pattern recognition
with Institute of Applied Mathematics, Academy of Mathematics and Systems Science, Chinese Academy of
Sciences, Beijing, China.
	
His current research interests include clustering, unsupervised feature learning, manifold learning,
metric learning and deep learning.
\end{IEEEbiography}

\begin{IEEEbiography}[{\includegraphics[width=1in,height=1.25in,clip,keepaspectratio]{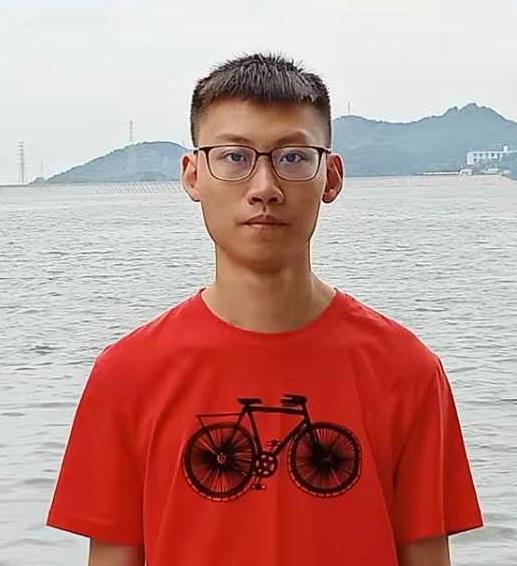}}]
{Ruihang~Xu} received BSc degree in mathematics and statistics from Xidian University, Xi'an, China, in 2018.
He is currently pursuing his PhD degree in machine learning with Institute of Applied Mathematics, Academy
of Mathematics and Systems Science, Chinese Academy of Sciences, Beijing, China.
His current research interests include deep learning and image processing.
\end{IEEEbiography}

\begin{IEEEbiography}[{\includegraphics[width=1in,height=1.25in,clip,keepaspectratio]{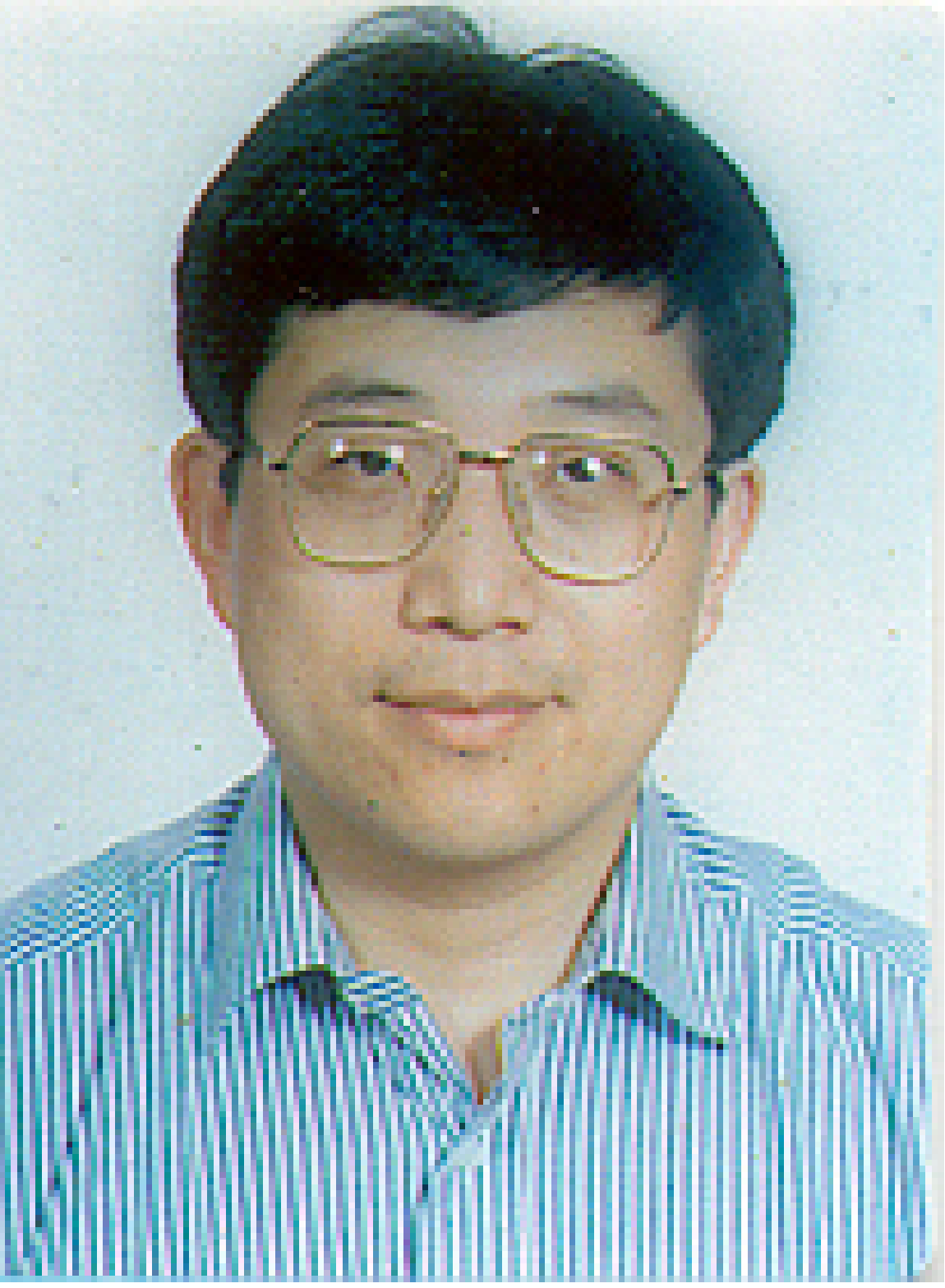}}]
{Bo~Zhang} (M'10) received BSc degree in mathematics from Shandong University, Jinan, China, MSc degree
in mathematics from Xi'an Jiaotong University, Xi'an, China, and PhD degree in applied mathematics
from the University of Strathclyde, Glasgow, UK, in 1983, 1985, and 1992, respectively.
	
After being a postdoc at Keele University, UK and a Research Fellow at Brunel University, UK, from 1992 to 1997,
he joined Coventry University, Coventry, UK, in 1997, as a Senior Lecturer, where he was promoted to
Reader in Applied Mathematics in 2000 and to Professor of Applied Mathematics in 2003. He is currently a
Professor with Institute of Applied Mathematics, Academy of Mathematics and Systems Science, Chinese Academy
of Sciences, Beijing, China. His current research interests include direct and inverse scattering problems,
radar and sonar imaging, machine learning, and data mining.
He is currently an Associate Editor of the IEEE TRANSACTION ON CYBERNETICS, and Applicable Analysis.
\end{IEEEbiography}

\begin{IEEEbiography}[{\includegraphics[width=1in,height=1.25in,clip,keepaspectratio]{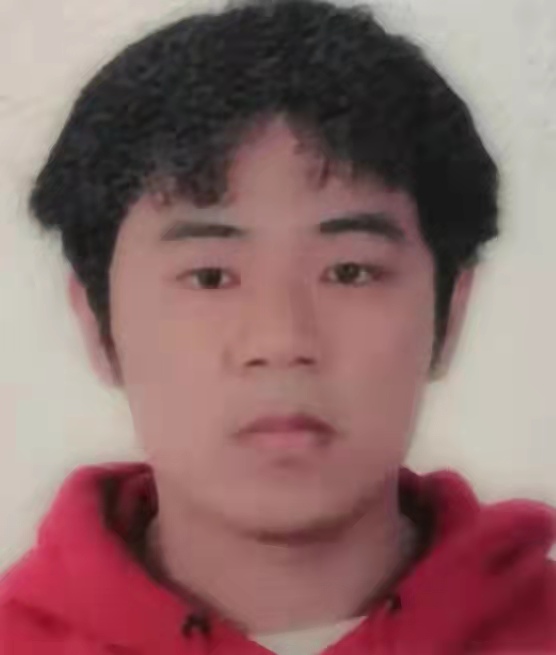}}]
{Ruinan~Jin} received the B.S.c. degree in Schiffsmotor from Wuhan University of Technology, China, in 2017.
He is currently pursuing the M.Sc degree in machine learning and pattern recognition with Academy of
Mathematics and Systems Science, Chinese Academy of Sciences, Beijing, China. His current research
interests include  stochastic optimization,
unsupervised feature learning, deep learning theory, causal discovery.
\end{IEEEbiography}

\end{document}